\pdfoutput=1

\documentclass[11pt]{article}

\usepackage[]{acl}

\usepackage{times}
\usepackage{latexsym}

\usepackage[T1]{fontenc}

\usepackage[utf8]{inputenc}

\usepackage{microtype}

\usepackage{inconsolata}

\usepackage{graphicx}

\usepackage{booktabs, tabularx, multirow, multicol, makecell, colortbl}

\usepackage{enumitem}

\usepackage{bbding}

\usepackage{amsmath, amssymb}

\usepackage{xspace}
\newcommand{\benchname}{CODIS\xspace}

\title{\benchname: Benchmarking Context-Dependent Visual Comprehension \\ for Multimodal Large Language Models}

\author{
\textbf{Fuwen Luo}$^{1*}$, \textbf{Chi Chen}$^{1*}$, \textbf{Zihao Wan}$^{1}$, \textbf{Zhaolu Kang}$^{6}$, \textbf{Qidong Yan}$^{5}$, \textbf{Yingjie Li}$^{5}$, \\\textbf{Xiaolong Wang}$^{1}$, \textbf{Siyu Wang}$^{2}$, \textbf{Ziyue Wang}$^{1}$, \textbf{Xiaoyue Mi}$^{7}$, \\\textbf{Peng Li}$^{2,3\dagger}$, \textbf{Ning Ma}$^{5}$, \textbf{Maosong Sun}$^{1\dagger}$, \textbf{Yang Liu}$^{1,2,3,4}$ \\
$^{1}$Dept. of Comp. Sci. \& Tech., Institute for AI, Tsinghua University, Beijing, China \\
$^{2}$Institute for AI Industry Research (AIR), Tsinghua University, Beijing, China \\
$^{3}$ Shanghai Artificial Intelligence Laboratory, Shanghai, China, \\
$^{4}$ Jiangsu Collaborative Innovation Center for Language Competence, Jiangsu, China \\
$^{5}$Key Laboratory of Linguistic and Cultural Computing Ministry of Education, \\ Northwest Minzu University, China \\
$^{6}$College of Software, Jilin University, China \\
$^{7}$Institute of Computing Technology, Chinese Academy of Sciences \\
\texttt{\{lfw23,chenchi19\}@mails.tsinghua.edu.cn} \\
\texttt{lipeng@air.tsinghua.edu.cn, sms@tsinghua.edu.cn} \\
}

\begin{document}
\maketitle
\renewcommand{\thefootnote}{\fnsymbol{footnote}}
\footnotetext[1]{Equal Contribution.} 
\footnotetext[2]{Corresponding Authors.}
\renewcommand{\thefootnote}{\arabic{footnote}}
\begin{abstract}
Multimodal large language models (MLLMs) have demonstrated promising results in a variety of tasks that combine vision and language. As these models become more integral to research and applications, conducting comprehensive evaluations of their capabilities has grown increasingly important. However, most existing benchmarks fail to consider that, in certain situations, images need to be interpreted within a broader context. In this work, we introduce a new benchmark, named as \benchname, designed to assess the ability of models to use context provided in free-form text to enhance visual comprehension. Our findings indicate that MLLMs consistently fall short of human performance on this benchmark. Further analysis confirms that these models struggle to effectively extract and utilize contextual information to improve their understanding of images. This underscores the pressing need to enhance the ability of MLLMs to comprehend visuals in a context-dependent manner. View our project website at \url{https://thunlp-mt.github.io/CODIS}.
\end{abstract}

\section{Introduction}

Recent years have witnessed a rapid advancement in multimodal large language models (MLLMs, \citealt{openai2023gpt4vision, gemini2023gemini, liu2023improved, dai2023instructblip}). They have achieved remarkable results on various downstream tasks, such as image captioning~\citep{luo2023semantic, wang2023controllable, chen2015microsoft}, visual question answering~\citep{shao2023prompting, liu2023cross, antol2015vqa} and visual reasoning~\citep{gupta2023visual, chen2023see, zellers2019recognition}. As the performance of MLLMs continues to improve, the comprehensive assessment of their capabilities becomes increasingly important.

\begin{figure}[t]
    \centering
    \includegraphics[width=0.45\textwidth]{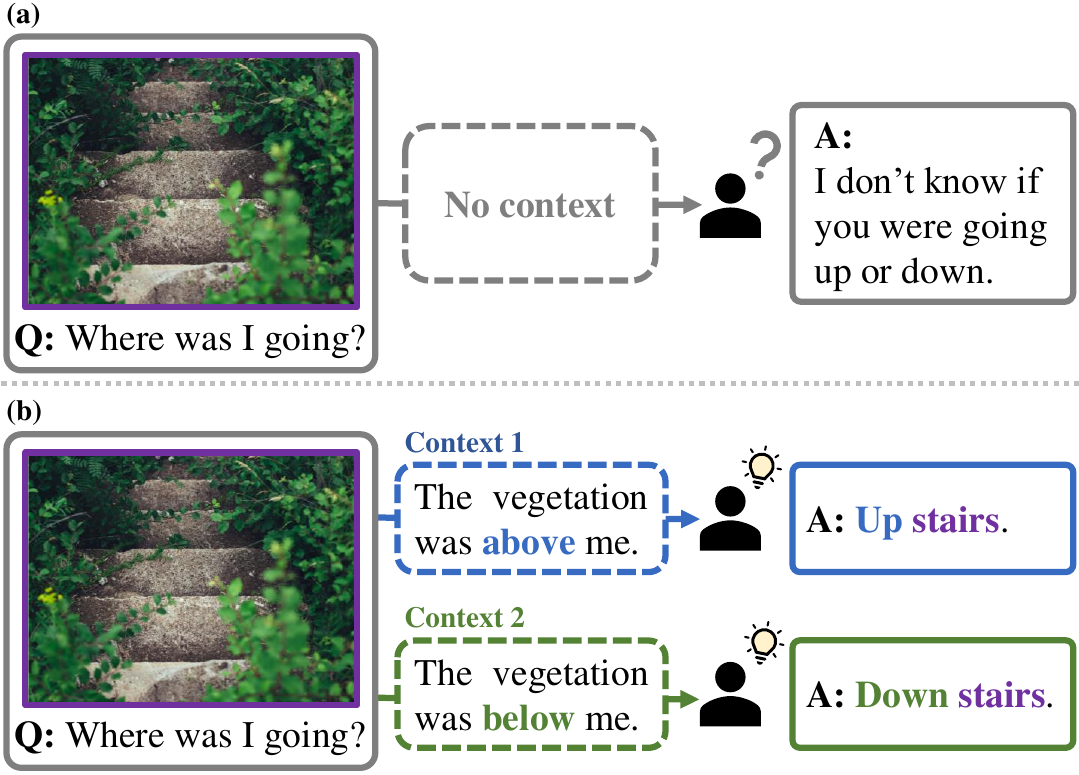}
    \caption{Interpretation of images can be significantly influenced by contextual information. In this instance, the determination of whether the photographer was ascending or descending a staircase remains ambiguous without supplementary context (a). However, when additional information is provided, indicating the position of the greenery relative to the observer, the direction of movement of the observer becomes clear (b). In the responses, words originating from the image and the two pieces of context are highlighted in \textcolor[RGB]{113,49,160}{purple}, \textcolor[RGB]{55,107,193}{blue}, and \textcolor[RGB]{84,130,53}{green}, respectively.}
    \label{fig:introduction}
\end{figure}

\begin{table*}[t]
\centering
\small
\setlength{\tabcolsep}{2pt}
\resizebox{0.98\textwidth}{!}{
\begin{tabular}{lccccc}
\toprule
\textbf{Benchmark} & \textbf{Purpose} & \textbf{Ambiguity} & \textbf{Context} & \textbf{Answer Type} & \textbf{Evaluator} \\
\midrule
MME~\citep{fu2023mme} & Comprehensive Evaluation & \color{purple}{\XSolidBrush} & \color{purple}{\XSolidBrush} & Yes/No & Metrics \\
MMBench~\citep{liu2023mmbench} & Comprehensive Evaluation & \color{purple}{\XSolidBrush} & \color{purple}{\XSolidBrush} & Multi-choice & GPT \\
SEED-Bench~\citep{li2023seed} & Comprehensive Evaluation & \color{purple}{\XSolidBrush} & \color{purple}{\XSolidBrush} & Multi-choice & Metrics \\\midrule
VisDial~\citep{das2017visual} & Visual Dialog & \color{purple}{\XSolidBrush} & \color{teal}{\Checkmark} & Open-ended & Metrics \\
MMDialog~\citep{feng2022mmdialog} & Visual Dialog & \color{purple}{\XSolidBrush} & \color{teal}{\Checkmark} & Open-ended & Metrics \\\midrule
HallusionBench~\citep{guan2023hallusionbench} & Visual Hallucination & \color{teal}{\Checkmark} & \color{purple}{\XSolidBrush} & Yes/No & Metrics \\
Bingo~\citep{cui2023holistic} & Visual Hallucination & \color{teal}{\Checkmark} & \color{purple}{\XSolidBrush} & Open-ended & Human \\
SQUID-E~\citep{sanders2022ambiguous} & Visual Ambiguity & \color{teal}{\Checkmark} & \color{purple}{\XSolidBrush} & Multi-choice & Metrics \\
DCIC~\citep{schmarje2022one} & Visual Ambiguity & \color{teal}{\Checkmark} & \color{purple}{\XSolidBrush} & Multi-choice & Metrics \\\midrule
\textbf{\benchname} (Ours) & Image-context Comprehension & \color{teal}{\Checkmark} & \color{teal}{\Checkmark} & Close \& Open-ended & Human / GPT \\\bottomrule
\end{tabular}}
\caption{Comparison of our proposed \benchname with recent vision-language benchmarks.}
\label{tab:other_benchmarks}
\end{table*}

Previous research highlights a fascinating and vital human ability: understanding visual elements within broader contexts~\citep{baldassano2016two}. Consider the example in Figure~\ref{fig:introduction}, where simply looking at an image does not reveal whether the stairs go up or down. However, introducing the context of whether the greenery is above or below the viewpoint of the photographer enables humans to clarify the ambiguity and accurately interpret the image. In this study, we argue that MLLMs should also be evaluated from the perspective of context-dependent visual comprehension. From a scientific perspective, it is important to understand the extent to which MLLMs can leverage context to enhance their visual comprehension. Practically speaking, as MLLMs are increasingly applied in real-world scenarios, assessing their reliability in interpreting visual data within context is essential for ensuring they provide accurate responses.

However, the capability for context-dependent visual understanding has not been fully assessed by current benchmarks for MLLMs. Table~\ref{tab:other_benchmarks} summarizes recent benchmarks for MLLMs. Most of these benchmarks~\citep{fu2023mme, liu2023mmbench, li2023seed, guan2023hallusionbench, cui2023holistic, sanders2022ambiguous, schmarje2022one} do not pair images with additional context. Only two benchmarks, namely VisDial~\citep{das2017visual} and MMDialog~\citep{feng2022mmdialog}, include extra context to help in conversation with humans rather than to clarify the meaning of images as shown in Figure~\ref{fig:introduction}. This limitation means these benchmarks are not fully capable of testing the ability of MLLMs to understand images in a context-dependent manner.

To address this challenge, we introduce a new benchmark, named as \benchname, designed to evaluate the capability of MLLMs in \textbf{CO}ntext-\textbf{D}ependent \textbf{I}mage di\textbf{S}ambiguation. \benchname utilizes the visual question answering (VQA,~\citealt{antol2015vqa}) format for this purpose, as shown in Figure~\ref{fig:introduction}(b). It stands out from existing benchmarks in three main aspects: first, each image in \benchname contains inherent ambiguity that can only be resolved with additional context; second, the questions are deliberately designed to highlight these ambiguities, requiring external context for accurate interpretation; third, for every image-question pair, we provide two contexts in a free-form text format. These contexts are subtly different yet lead to different interpretations of the image and, consequently, different answers. We have carefully curated all images, questions, and contexts by hand to maintain a high standard of quality and diversity. Our assessment of 14 widely-used MLLMs with \benchname indicates that the performance of these models in understanding context-dependent visuals significantly falls short of human capabilities. Further analysis demonstrates that the models struggle with identifying crucial contextual cues and extracting relevant visual features.

To summarize, our contributions are three-fold:

\begin{itemize}
    \setlength{\itemsep}{0pt}
    \item We highlight the significance of context-dependent visual comprehension abilities for MLLMs.
    \item We introduce the \benchname benchmark to assess capabilities of MLLMs on context-dependent visual comprehension.
    \item Through our analysis, we uncover the deficiencies in MLLMs regarding context information extraction and visual information extraction, underscoring the immense potential for enhancement in the realm of context-dependent visual comprehension.
\end{itemize}

\section{Related Work}

\subsection{Context in Visual Tasks}

Previous works emphasize the importance of context in visual tasks~\citep{wang2023context, vo2021meaning, goh2004cortical, bar2003cortical}, exploring utilizing context in scene graph generation~\citep{cong2023reltr, zheng2023prototype, zhu2022scene, yang2018graph}, object detection~\citep{zou2023object, diwan2023object, xiao2023tiny, chen2023diffusiondet}, image inpainting~\citep{zhang2023image, sargsyan2023mi, xiang2023deep}, etc. However, they mainly focus on visual context within images, e.g., relationships between objects, environment and background. External contexts, such as free-form texts, have also been studied. Previous works have shown their effectiveness in tasks such as floorplan reconstruction~\citep{purushwalkam2021audio}, and visual dialog~\citep{das2017visual, feng2022mmdialog}. The ability of multimodal in-context learning~\citep{li2023mimic, zhao2023mmicl, sun2023generative, li2023fine} further demonstrates the significance of external contexts. However, in these works, context do not affect the interpretation of images, which is the main focus of our work. We strengthen the impact of context on visual comprehension.

\begin{figure*}[t]
    \centering
    \includegraphics[width=1.0\textwidth]{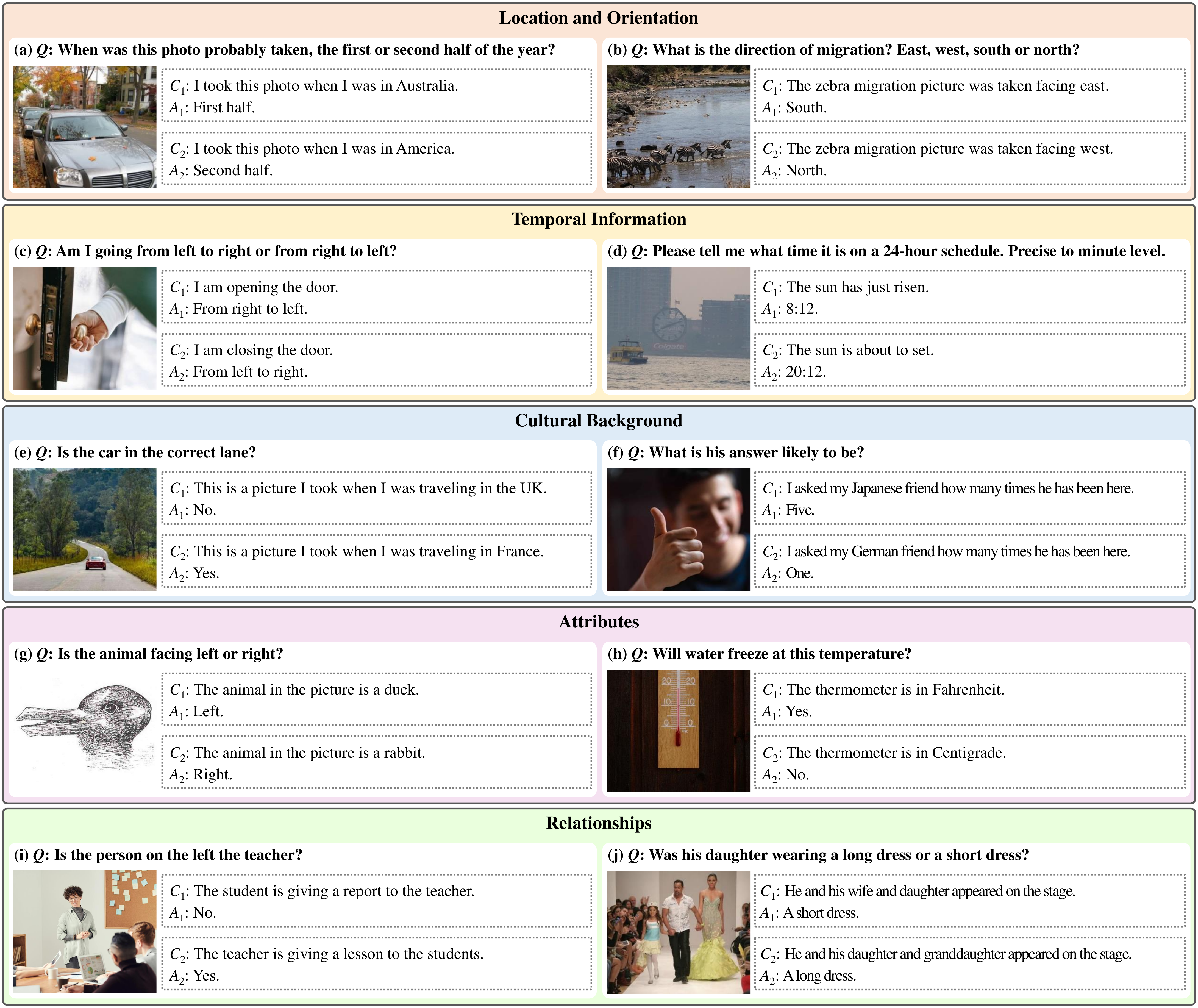}
    \caption{Taxonomy of our benchmark. We show two cases for each category. In each case, we show an image and a question, along with two pieces of different context and their corresponding answers. We use Q to denote questions, C for contexts and A for answers.}
    \label{fig:taxonomy}
\end{figure*}

\subsection{Visual Ambiguities}

Visual ambiguities arises from absence of visual information or other noises~\citep{denison2018humans}. It is an important phenomenon in human visual cognition. Previous works investigate visual ambiguities caused by optical illusions~\citep{guan2023hallusionbench, cui2023holistic, fu2023challenger}. Images which may cause illusions are leveraged to evaluate visual hallucinations in MLLMs. Other works explore ambiguities in visual event classification~\citep{sanders2022ambiguous, schmarje2022one, rajeswar2022multi}, focusing on model robustness against ambiguous images. Different from these works, we assess capability of MLLMs to disambiguate images instead of just recognizing ambiguities.

\section{\benchname}
\benchname is proposed for evaluating the capability of MLLMs in context-dependent image disambiguation. Figure~\ref{fig:taxonomy} presents several examples from our benchmark, highlighting the diversity of contexts covered. In this section, we first elaborate our taxonomy of context. Then, we delve into the instruction design and evaluation method of \benchname. Finally, we introduce procedures of data collection.

\subsection{Taxonomy of Context}

Given the extensive and varied nature of context information, it is challenging to catalog all forms of context comprehensively. Nevertheless, in our effort to establish a pioneering benchmark for context-dependent image disambiguation, we aim to cover a wide range of diverse and illustrative scenarios. Drawing inspiration from the types of information people require to understand the visual contents of an image, we have identified five representative types of context. These include three types of global context that pertain to the overall scene—namely, the global background, which encompasses location and orientation, temporal information, and cultural background. Additionally, we focus on two types of local context related to objects within the scene, specifically the attributes of objects and the relationships between people. We present examples in Figure~\ref{fig:taxonomy} and provide an explanation of our classification as follows.

\textbf{Location and orientation.} Interpretations of scenes is closely related to where they happen. As Figure~\ref{fig:taxonomy}(a) shows, the scene of falling leaves indicates different times in a year when they happen in different countries. Orientation also plays an important role in image understanding. In Figure~\ref{fig:taxonomy}(b), direction which the photographer is facing can help determine direction in which the herd is moving.

\textbf{Temporal information.} Understanding the timing and sequence of events is crucial when we understand a scene. However, an image can only provide us with static information. Thus, context of temporal information is often crucial, and can help better restore the original appearance of events. Figure~\ref{fig:taxonomy}(c) shows an example of context which indicates dynamic changes of things and events, like opening or closing the door, while context of Figure~\ref{fig:taxonomy}(d) tells when pictures were taken.
\footnote{One might argue that introducing video is the most direct method to resolve temporal ambiguities. However, videos are not always available in practice. By analogy to humans, a sufficiently capable MLLM should also be able to resolve such ambiguities without relying on videos. Therefore, we believe it is necessary to include this category of contexts in our benchmark.}

\textbf{Cultural background.} Cultural background has a profound impact on the way people think and act. Meaning of an action may have significant differences across different nations and cultures sometimes. Awareness of cultural background is crucial for understanding images. We employ cases covering various aspects of differences between cultures, such as laws, regulations, customs and traditions. See Figure~\ref{fig:taxonomy}(e) and Figure~\ref{fig:taxonomy}(f) for examples.

\textbf{Attributes.} To comprehensively understand objects in a scene, it is not enough to solely rely on visual appearances we see. Non-visible attributes are also important. We consider two types of attributes, including \textit{physical attributes}, such as structure, size, temperature, smell and touch, as Figure~\ref{fig:taxonomy}(g) shows, and \textit{abstract attributes}, such as function and mechanism, as Figure~\ref{fig:taxonomy}(h) shows.

\textbf{Relationships.} We communicate with people in different ways based on our relationships with them. When the same event happens on people with different relationships, its meaning varies. It is important to understand relationships between people in visual comprehension. As Figure~\ref{fig:taxonomy}(i) and Figure~\ref{fig:taxonomy}(j) shows, relationships between people affect our judgement on their roles and identities.

\begin{table*}[t]
\centering
\small
\begin{tabular}{lcccc}
\toprule
\textbf{Model} & \textbf{Parameters} & \textbf{Image Encoder} & \textbf{LLM} & \textbf{Alignment Module} \\\midrule
GPT-4V~\citep{openai2023gpt4vision} & \multirow{2}{*}{-} & - & - & - \\
Gemini~\citep{gemini2023gemini} & & - & - & - \\\midrule
LLaVA-1.5-13B~\citep{liu2023improved} & \multirow{3}{*}{\textasciitilde{}13B} & CLIP\textsubscript{ViT-L}-336px  & LLaMA-2-13B & MLP \\
BLIP-2-11B~\citep{li2023blip} & & EVA-CLIP\textsubscript{ViT-G} & Flan-T5-xxl & Q-Former \\
InstructBLIP-13B~\citep{dai2023instructblip} & & EVA-CLIP\textsubscript{ViT-G} & Vicuna-13B & Q-Former \\\midrule
mPLUG-Owl-2~\citep{ye2023mplugowl2} & \multirow{9}{*}{\textasciitilde{}7B} & CLIP\textsubscript{ViT-L} & LLaMA-2-7B & Attention \\
MiniGPT4-7B~\citep{zhu2023minigpt} & & EVA-CLIP\textsubscript{ViT-G} & Vicuna-7B & Linear \\
LLaVA-1.5-7B~\citep{liu2023improved} & & CLIP\textsubscript{ViT-L}-336px  & LLaMA-2-7B & MLP \\
InstructBLIP-7B~\citep{dai2023instructblip} & & EVA-CLIP\textsubscript{ViT-G} & Vicuna-7B & Q-Former \\
Otter-7B~\citep{li2023mimic} & & CLIP\textsubscript{ViT-L} & MPT-7B & Attention \\
LLaVA-7B~\citep{liu2023visual} & & CLIP\textsubscript{ViT-L} & Vicuna-7B & Linear \\
Qwen-VL-Chat~\citep{bai2023qwen} & & OpenCLIP\textsubscript{ViT-bigG} & Qwen-7B & Attention \\
OpenFlamingo-7B~\citep{awadalla2023openflamingo} & & CLIP\textsubscript{ViT-L} & MPT-7B & Attention \\
BLIP-2-6.7B~\citep{li2023blip} & & EVA-CLIP\textsubscript{ViT-G} & OPT-6.7B & Q-Former \\\bottomrule
\end{tabular}
\caption{API-based and open-source MLLMs selected for evaluation.}
\label{tab:models}
\end{table*}

\subsection{Instruction Design}

To prevent models from guessing the correct answers without fully understanding context, we organize our dataset in pairs. Each pair contains two queries $ (I,Q,C_1) $ and $ (I,Q,C_2) $. The queries have identical image $ I $ and question $ Q $, but have two pieces of different context $ C_1 $ and $ C_2 $. We give MLLMs two queries separately, and get model outputs $ O_1 $ and $ O_2 $. Take Figure~\ref{fig:taxonomy}(a) as an example. In the first query, we provide a picture of falling leaves and indicate that the photo was taken in Australia. In the second query, we give different context that the photo was taken in America. We expect to get different outputs from models that the photo was taken in the first or the second half of the year.

\subsection{Evaluation Method}

We use $ O_1 $ and $ O_2 $ to denote model outputs of a pair of queries, and $ A_1 $ and $ A_2 $ to denote the groundtruth answers. The evaluation of model outputs can be represented as follows:
$$
\mathrm{Eval}(O_i)=
\begin{cases}
1 & \mathrm{if}~O_i~\mathrm{matches}~A_i \\
0 & \mathrm{otherwise}
\end{cases},i=1,2\text{.}
$$

Following \citet{fu2023mme}, we leverage two metrics, pair-wise accuracy $ \mathrm{Acc}_p $ and query-wise accuracy $ \mathrm{Acc}_q $ for our evaluation metrics, which are calculated as follows:
$$
\begin{aligned}
\mathrm{Acc}_p&=\frac{1}{n_p}\sum_{i=1}^{n_p}(\mathrm{Eval}(O_{i1})\times\mathrm{Eval}(O_{i2}))\text{,}\\
\mathrm{Acc}_q&=\frac{1}{2n_p}\sum_{i=1}^{n_p}(\mathrm{Eval}(O_{i1})+\mathrm{Eval}(O_{i2}))\text{,}
\end{aligned}
$$
where $ n_p $ is number of data pairs. For $ \mathrm{Acc}_p $, models score only if their answers to a pair of queries are both correct. For $ \mathrm{Acc}_q $, models score for each single query they answer correctly.

\subsection{Data Collection}

In this section, we introduce the process of constructing our benchmark. Our data collection process includes three steps.

\textbf{Image collection.} We manually collect images that contain ambiguities which can only be resolved with external contexts. The majority of these images are real-scene images from the publicly available dataset ShareGPT4V~\citep{chen2023sharegpt4v} and the Internet, while the remainder are created manually. The primary challenge in collecting images was that many appeared ambiguous at first glance but were actually unambiguous upon closer examination. Therefore, each image had to be carefully inspected and excluded from the collection if it was found to be unambiguous. In total, we collected 377 images.

\begin{table*}[t]
\centering
\small
\resizebox{0.98\textwidth}{!}{
\begin{tabular}{l|cccccccccc|cc|c}
\toprule
\multirow{2}{*}{\textbf{Model}} & \multicolumn{2}{c}{\textbf{Loc \& Ori}} & \multicolumn{2}{c}{\textbf{Temporal}} & \multicolumn{2}{c}{\textbf{Cultural}} & \multicolumn{2}{c}{\textbf{Attributes}} & \multicolumn{2}{c|}{\textbf{Relationships}} & \multicolumn{2}{c|}{\textbf{Overall}} & \multirow{2}{*}{\textbf{\makecell[c]{Context \\ Awareness}}} \\
& \textbf{Acc\textsubscript{\textit{p}}} & \textbf{Acc\textsubscript{\textit{q}}} & \textbf{Acc\textsubscript{\textit{p}}} & \textbf{Acc\textsubscript{\textit{q}}} & \textbf{Acc\textsubscript{\textit{p}}} & \textbf{Acc\textsubscript{\textit{q}}} & \textbf{Acc\textsubscript{\textit{p}}} & \textbf{Acc\textsubscript{\textit{q}}} & \textbf{Acc\textsubscript{\textit{p}}} & \textbf{Acc\textsubscript{\textit{q}}} & \textbf{Acc\textsubscript{\textit{p}}} & \textbf{Acc\textsubscript{\textit{q}}} & \\\midrule\noalign{\vskip -3pt}
\rowcolor{gray!20}\multicolumn{14}{c}{\textbf{Human}} \\\noalign{\vskip -2pt}\midrule
Human & 85.2 & 86.1 & 90.9 & 92.8 & 72.8 & 76.4 & 87.2 & 88.4 & 89.6 & 90.0 & 86.2 & 87.7 & 97.3 \\\midrule\noalign{\vskip -3pt}
\rowcolor{gray!20}\multicolumn{14}{c}{\textbf{API-based Models}} \\\noalign{\vskip -2pt}\midrule
GPT-4V & 33.3 & 54.2 & 28.4 & 52.1 & 25.5 & 60.6 & 26.7 & 54.7 & 51.9 & 70.2 & 32.3 & 56.9 & 54.7 \\
Gemini & 21.4 & 49.4 & 29.5 & 51.1 & 21.3 & 56.4 & 24.0 & 52.0 & 34.6 & 58.7 & 26.1 & 52.7 & 43.6 \\\midrule\noalign{\vskip -3pt}
\rowcolor{gray!20}\multicolumn{14}{c}{\textbf{Open-source \textasciitilde{}13B Models}} \\\noalign{\vskip -2pt}\midrule
LLaVA-1.5-13B & \hphantom{0}\textbf{6.0} & \textbf{41.1} & \hphantom{0}\underline{4.2} & \underline{44.7} & \textbf{10.6} & \textbf{50.0} & \textbf{14.7} & \textbf{51.3} & \textbf{13.5} & \textbf{54.8} & \hphantom{0}\textbf{9.1} & \textbf{47.5} & 19.3 \\
BLIP-2-11B & \hphantom{0}\textbf{6.0} & 32.7 & \hphantom{0}\textbf{8.4} & \textbf{45.8} & \hphantom{0}\underline{4.3} & 35.1 & \hphantom{0}\underline{6.7} & 42.0 & \underline{11.5} & \underline{51.9} & \hphantom{0}\underline{7.4} & 41.4 & 31.4 \\
InstructBLIP-13B & \hphantom{0}\textbf{6.0} & \underline{39.3} & \hphantom{0}2.1 & 41.6 & \hphantom{0}\underline{4.3} & \textbf{50.0} & \hphantom{0}4.0 & \underline{44.7} & \hphantom{0}7.7 & 51.0 & \hphantom{0}4.5 & \underline{44.2} & 14.1 \\\midrule\noalign{\vskip -3pt}
\rowcolor{gray!20}\multicolumn{14}{c}{\textbf{Open-source \textasciitilde{}7B Models}} \\\noalign{\vskip -2pt}\midrule
mPLUG-Owl-2 & \textbf{13.1} & \underline{42.3} & \hphantom{0}\textbf{9.5} & 41.6 & \hphantom{0}\textbf{6.4} & 42.6 & \textbf{12.0} & \underline{44.7} & \textbf{19.2} & \underline{51.9} & \textbf{11.9} & \underline{44.1} & 31.7 \\
MiniGPT4-7B & 10.7 & 36.3 & \hphantom{0}3.2 & 34.2 & \hphantom{0}0.0 & 27.7 & \textbf{12.0} & 35.3 & \underline{13.5} & 47.1 & \hphantom{0}\underline{7.9} & 36.0 & 36.3 \\
LLaVA-1.5-7B & \underline{11.9} & \textbf{42.9} & \hphantom{0}5.3 & \underline{44.7} & \hphantom{0}\underline{4.3} & 43.6 & \hphantom{0}9.3 & 39.3 & \hphantom{0}7.7 & 47.1 & \hphantom{0}\underline{7.9} & 43.3 & 21.5 \\
InstructBLIP-7B & \hphantom{0}1.2 & 33.3 & \hphantom{0}\underline{7.4} & \textbf{45.8} & \hphantom{0}0.0 & \underline{46.8} & \hphantom{0}4.0 & 43.3 & 11.5 & 48.1 & \hphantom{0}4.8 & 42.8 & 16.7 \\
Otter-7B & \hphantom{0}2.4 & 32.7 & \hphantom{0}5.3 & 41.1 & \hphantom{0}\underline{4.3} & 28.7 & \hphantom{0}0.0 & 26.0 & \hphantom{0}5.8 & 40.4 & \hphantom{0}3.4 & 34.1 & 19.3 \\
LLaVA-7B & \hphantom{0}2.4 & 30.4 & \hphantom{0}6.3 & 34.2 & \hphantom{0}0.0 & 25.5 & \hphantom{0}1.3 & 34.0 & \hphantom{0}5.8 & 41.3 & \hphantom{0}3.4 & 33.1 & 17.2 \\
Qwen-VL-Chat & \hphantom{0}3.6 & 23.8 & \hphantom{0}3.2 & 24.7 & \hphantom{0}0.0 & 24.5 & \hphantom{0}1.3 & 32.0 & \hphantom{0}9.6 & 34.6 & \hphantom{0}3.4 & 27.5 & 26.3 \\
OpenFlamingo-7B & \hphantom{0}2.4 & 40.5 & \hphantom{0}2.1 & 38.9 & \hphantom{0}0.0 & 27.7 & \hphantom{0}5.3 & 36.0 & \hphantom{0}5.8 & 47.1 & \hphantom{0}3.1 & 38.4 & 15.6 \\
BLIP-2-6.7B & \hphantom{0}0.0 & 41.1 & \hphantom{0}1.1 & \underline{44.7} & \hphantom{0}2.1 & \textbf{48.9} & \hphantom{0}2.7 & \textbf{46.0} & \hphantom{0}7.7 & \textbf{53.8} & \hphantom{0}2.3 & \textbf{46.0} & \hphantom{0}6.5 \\\bottomrule
\end{tabular}}
\caption{Results of MLLMs on \benchname. All the model outputs are assessed by human. Consistency of human evaluation is investigated in Appendix~\ref{sec:appendix_consistency_human_eval}.}
\label{tab:results}
\end{table*}

\textbf{Design of questions, context and answers.} For each collected image, we manually write questions, context and answers for it. The data are compiled with following rules:

(1) Questions are targeted at the ambiguous parts of the images. They can not be answered if no additional context is provided due to ambiguities in the images.

(2) For each image-question pair, two unique contexts with minor differences that lead to different interpretations of the image need to be created. As a result, these interpretations yield different answers to the question. Furthermore, we require that an answer can only be derived when both the image and its corresponding context are available. This means that neither the image nor the context alone is sufficient to provide an answer. For instance, in Figure~\ref{fig:taxonomy}(a), the photograph of falling leaves, when combined with the knowledge that the photograph was taken in Australia, allows us to conclude that it depicts autumn in the southern hemisphere, occurring in the first half of the year. In isolation, neither the image nor the context is adequate to reach this conclusion.

(3) To enhance the evaluation process, MLLMs are permitted to respond in free-form text. However, to maintain a balance between the ease of evaluation and ensuring objectivity, we require the answers to be relatively objective. For example, despite the lack of a strict format for the responses to the scenarios shown in Figure~\ref{fig:taxonomy}, we provide clear options for models to choose in most of the cases. This approach guarantees that the responses stay objective and are easy to evaluate.

\begin{figure}[t]
    \centering
    \includegraphics[width=0.45\textwidth]{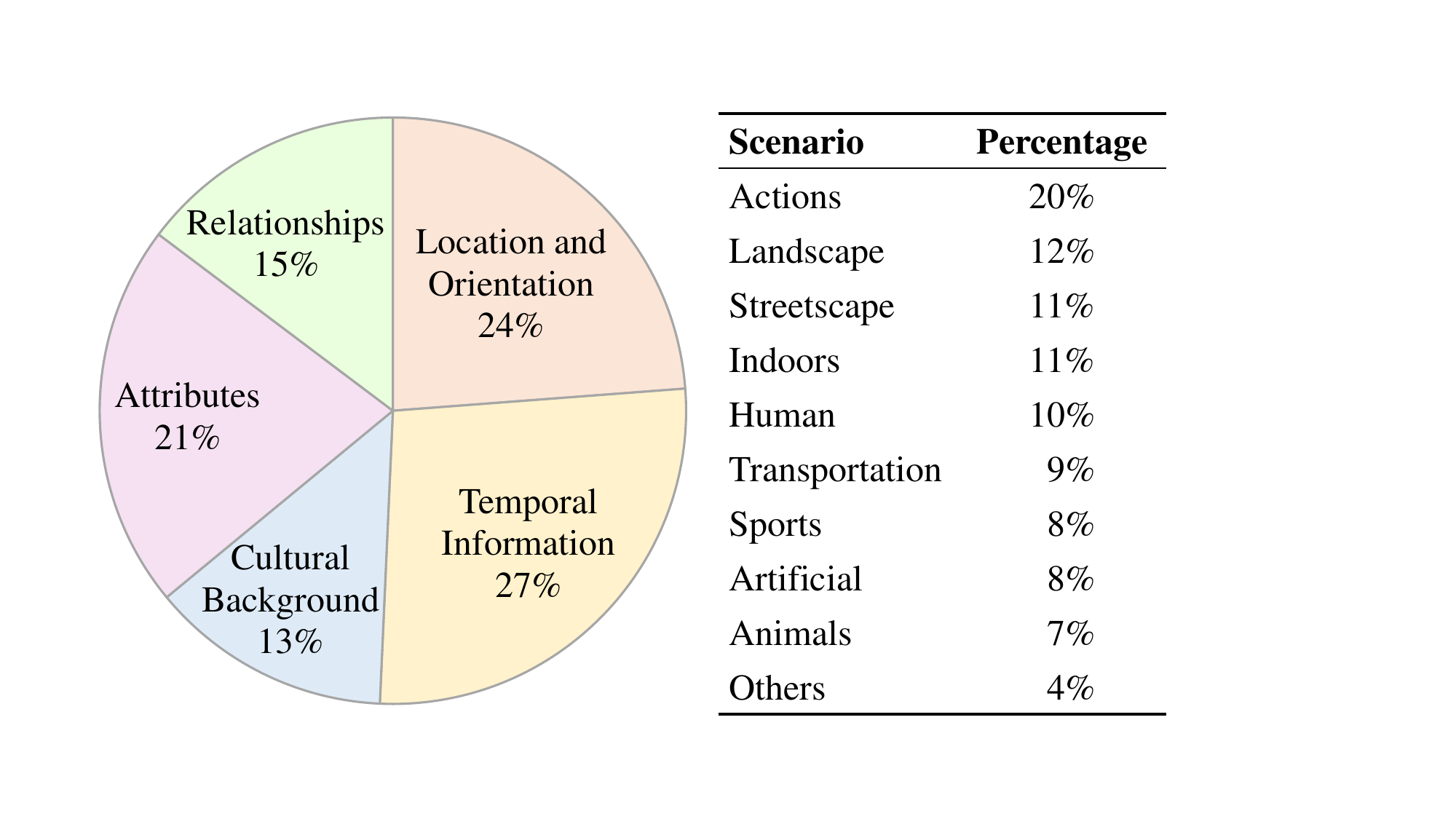}
    \caption{Distribution of five categories (left) and scenarios (right) of our \benchname benchmark.}
    \label{fig:categories}
\end{figure}

\textbf{Data verification.} Five annotators participated in the data collection process. To ensure the quality of our dataset, each submission by an annotator was cross-checked by the other four. Data was retained only if it (1) was correct, (2) differed from existing data to a significant extent, thereby maintaining high diversity, and (3) conformed to all the above criteria. If these conditions were not met, the annotator had to revise the data.

Ultimately, we retained 216 images and successfully collected a total of 706 queries, spanning five categories and encompassing a wide range of scenarios. We had to exclude 161 images because we were unable to collect eligibility questions or contexts for them. The distribution of categories and scenarios is illustrated in Figure~\ref{fig:categories}.

\section{Evaluation}

\subsection{Models}

As shown in Table~\ref{tab:models}, we perform evaluation on 14 popular MLLMs, which are divided into three groups: (1) \textbf{API-based models:} GPT-4V~\citep{openai2023gpt4vision} and Gemini~\citep{gemini2023gemini}; (2) \textbf{Open-source \textasciitilde{}7B models:} mPLUG-Owl-2~\citep{ye2023mplugowl2}, MiniGPT-4-7B~\citep{zhu2023minigpt}, LLaVA-1.5-7B~\citep{liu2023improved}, InstructBLIP-7B~\citep{dai2023instructblip}, Otter-7B~\citep{li2023mimic}, LLaVA-7B~\citep{liu2023visual}, Qwen-VL~\citep{bai2023qwen}, OpenFlamingo~\citep{awadalla2023openflamingo} and BLIP-2-6.7B~\citep{li2023blip}; (3) \textbf{Open-source \textasciitilde{}13B models:} LLaVA-1.5-13B, BLIP-2-11B and InstructBLIP-13B.

\subsection{Results}\label{sec:results}

We manually assess the outputs of MLLMs on \benchname benchmark and report the results in Table~\ref{tab:results}. ``Human'' refers to the average performance of five independent people, and the detailed results and analysis are shown in Appendix~\ref{sec:appendix_human_test}. 

\textbf{Overall results.} From an overall perspective, human is far ahead of all the MLLMs in all the categories, indicating that MLLMs have tremendous potential of improvement on context-dependent visual comprehension. API-based models significantly outperform open-source models, and GPT-4V gets the highest score among all the MLLMs. In terms of open-source models, mPLUG-Owl-2 has the best performance on $ \mathrm{Acc}_p $ and LLaVA-1.5-13B has the best performance on $ \mathrm{Acc}_q $.

\begin{table}[t]
\centering
\small
\begin{tabular}{lc}
\toprule
\textbf{Model} & \textbf{Output Variability} \\\midrule
Human & 81.9 \\\midrule
GPT-4V & 58.6 \\
Gemini & 53.5 \\\midrule
mPLUG-Owl-2 & 36.8 \\
LLaVA-1.5-13B & 25.6 \\\bottomrule
\end{tabular}
\caption{Variations in model outputs when context is removed. See the calculation of output variability in Section~\ref{sec:results}.}
\label{tab:res_remove_context}
\end{table}

\begin{table}[t]
\centering
\small
\begin{tabular}{lcc}
\toprule
\textbf{Model} & \textbf{\makecell[c]{w/ Images \\ w/o Captions}} & \textbf{\makecell[c]{w/ Captions \\ w/o Images}} \\\midrule
GPT-4V & 35.1 & 51.0 \\
Gemini & 31.1 & 35.8 \\\midrule
mPLUG-Owl-2 & 10.6 & 16.6 \\
LLaVA-1.5-13B & 11.9 & 20.5 \\\bottomrule
\end{tabular}
\caption{Variations in $ \mathrm{Acc}_p $ when we replace the images with their captions.}
\label{tab:res_replace_image_with_caption}
\end{table}

\textbf{Disparity between $ \mathrm{Acc}_p $ and $ \mathrm{Acc}_q $.} We observe a large disparity between $ \mathrm{Acc}_p $ and $ \mathrm{Acc}_q $ of MLLMs, and open-source models have a larger disparity than API-based models. To better understand the reason of this phenomenon, we further investigate the metric of context awareness, which reflects model capability of recognizing different context and provide different responses accordingly. It is calculated as follows:
$$
\mathrm{ContextAwareness}=\frac{1}{n_p}\sum_{i=1}^{n_p}(O_{i1}\neq O_{i2})\text{,}
$$
where $ n_p $ is number of data pairs. \textbf{We manually assess inequality of model outputs on semantic level.} The results are shown in the last column of Table~\ref{tab:results}. The scores of all the MLLMs are lower than 60, while human score is 97.3, indicating that for many data pairs, MLLMs fail to recognize the difference in context and they provide semantically equal response to the two queries. Thus, for a significant number of cases, models only correctly answer one of the two queries, leading to high $ \mathrm{Acc}_q $ and low $ \mathrm{Acc}_p $ scores. Additionally, context awareness is positively correlated with $ \mathrm{Acc}_p $, and has no clear correlation with $ \mathrm{Acc}_q $. Therefore, $ \mathrm{Acc}_p $ is more possible to reflect model capability of context comprehension, and is more suitable as the primary metric for \benchname benchmark. We further investigate the capability of MLLMs on extracting information from context. By excluding context and providing only images and questions as inputs, we observe the metric of output variability, which reflects the variations in model outputs. It is calculated as follows:
$$
\mathrm{OutputVariability}=\frac{1}{n_q}\sum_{i=1}^{n_q}{(O_{nc}^i\neq O_{c}^i)}\text{,}
$$
where $ n_q $ is the total number of queries, $ O_{nc} $ and $ O_{c} $ are outputs when context is given and is not given. \textbf{Inequality of model outputs is assessed manually on semantic level.} We conduct our experiment on API-based models GPT-4V and Gemini, and mPLUG-Owl-2 and LLaVA-1.5-13B, which have the highest $ \mathrm{Acc}_p $ and $ \mathrm{Acc}_q $ among open-source models, respectively. Additionally, we report human results for reference. As shown in Table~\ref{tab:res_remove_context}, MLLMs score significantly lower than human, which means they are incapable to change their answers according to context they received. Their capability of context information extraction is needed to be improved. Examples of model responses when removing context are shown in Appendix~\ref{sec:appendix_case_remove_context}.

\textbf{Comparison between \textasciitilde{}7B and \textasciitilde{}13B models.} We observe from Table~\ref{tab:results} that scaling up from \textasciitilde{}7B to \textasciitilde{}13B does not necessarily lead to an improvement in model performance. For LLaVA-1.5, its \textasciitilde{}13B model significantly outperforms \textasciitilde{}7B model on both $ \mathrm{Acc}_p $ and $ \mathrm{Acc}_q $. However, for BLIP series models, $ \mathrm{Acc}_p $ of InstructBLIP and $ \mathrm{Acc}_q $ of BLIP-2 slightly drop when they scale up. Regrettably, significant differences exist between the two models, covering various aspects including data, training paradigms, and additional details. Consequently, we were unable to determine the primary causes of the differing behaviors observed in the two models. While intensive efforts are warranted to explore these causes, we believe that such an investigation exceeds the scope of this work.

\textbf{Model performance on different categories.} Among the five categories, most MLLMs perform best on relationships. For open-source MLLMs, it is possibly because their training data include dense image annotations~\citep{liu2023visual, liu2023improved, ye2023mplugowl2} such as Visual Genome~\citep{krishna2017visual}, or detailed captions~\citep{li2023blip, dai2023instructblip, zhu2023minigpt} which contain lots of data of relationships. We also observe that the score of MLLMs on cultural background is relatively low, indicating they are not sufficiently trained on culture-related data.

\section{Analyses}

\subsection{Visual Information Extraction}

Extracting information from images according to context is the foundation of context-dependent visual comprehension task. We explore capability of MLLMs on extracting visual information. We replace images with detailed captions and observe the changes in $ \mathrm{Acc}_p $. For some of the pictures are not easy to be described in natural language, we discard these images and only perform experiments on a subset of 96 images, 302 queries. We ensure that captions and context are sufficient for answering the questions. We conduct our experiment on GPT-4V, Gemini, mPLUG-Owl-2 and LLaVA-1.5-13B. The model outputs are assessed manually and results are shown in Table~\ref{tab:res_replace_image_with_caption}. For all the MLLMs, $ \mathrm{Acc}_p $ increases when the images are replaced with captions, indicating that compared to images, MLLMs are more skilled at extracting information from text with the help of context. Model capability to extract visual information according to context is needed to be improved. Examples of model responses when replacing images with captions are shown in Appendix~\ref{sec:appendix_case_replace_image_with_caption}.

\begin{table}[t]
\centering
\small
\resizebox{0.48\textwidth}{!}{
\begin{tabular}{lccc}
\toprule
\textbf{Model} & \textbf{\makecell[c]{up:\\down}} & \textbf{\makecell[c]{morning:\\afternoon}} & \textbf{\makecell[c]{appropriate:\\not appropriate}} \\\midrule
Groundtruth & 15:15 & 16:16 & 13:13 \\\midrule
GPT-4V & 14:12 & \hphantom{0}7:12 & 16:9\hphantom{0} \\
Gemini & 18:11 & \hphantom{0}5:21 & 22:2\hphantom{0} \\\midrule
mPLUG-Owl-2 & 23:6\hphantom{0} & \hphantom{0}9:15 & 23:1\hphantom{0} \\
LLaVA-1.5-13B & 21:8\hphantom{0} & 19:7\hphantom{0} & 22:4\hphantom{0} \\\bottomrule
\end{tabular}}
\caption{Bias in model outputs. We report frequencies of three pairs of specific words. Some outputs do not contain either of the two words, which makes total word frequencies in model outputs less than groundtruth.}
\label{tab:res_model_bias}
\end{table}

\begin{table}[t]
\centering
\small
\begin{tabular}{lccc}
\toprule
\multirow{3}{*}{\textbf{Model}} & \multicolumn{2}{c}{\textbf{Acc\textsubscript{\textit{p}}}} & \multirow{3}{*}{\textbf{Agreement}} \\
& \textbf{\makecell[c]{Human \\ Eval}} & \textbf{\makecell[c]{GPT-4 \\ Eval}} & \\\midrule
GPT-4V & 32.3 & 31.2 & \hphantom{0}97.6 \\
Gemini & 26.1 & 24.4 & \hphantom{0}97.7 \\\midrule
LLaVA-1.5-13B & \hphantom{0}9.1 & \hphantom{0}8.5 & \hphantom{0}97.6 \\
BLIP-2-11B & \hphantom{0}7.4 & \hphantom{0}7.9 & \hphantom{0}94.8 \\
InstructBLIP-13B & \hphantom{0}4.5 & \hphantom{0}4.5 & \hphantom{0}98.6 \\\midrule
mPLUG-Owl-2 & 11.9 & \hphantom{0}9.9 & \hphantom{0}96.7 \\
MiniGPT4-7B & \hphantom{0}7.9 & \hphantom{0}6.8 & \hphantom{0}91.8 \\
LLaVA-1.5-7B & \hphantom{0}7.9 & \hphantom{0}5.7 & \hphantom{0}94.3 \\
InstructBLIP-7B & \hphantom{0}4.8 & \hphantom{0}4.2 & \hphantom{0}98.7 \\
Otter-7B & \hphantom{0}3.4 & \hphantom{0}2.5 & \hphantom{0}94.3 \\
LLaVA-7B & \hphantom{0}3.4 & \hphantom{0}2.5 & \hphantom{0}92.8 \\
Qwen-VL-Chat & \hphantom{0}3.4 & \hphantom{0}3.4 & \hphantom{0}97.7 \\
OpenFlamingo-7B & \hphantom{0}3.1 & \hphantom{0}3.1 & 100.0 \\
BLIP-2-6.7B & \hphantom{0}2.3 & \hphantom{0}2.5 & \hphantom{0}98.1 \\\bottomrule
\end{tabular}
\caption{Comparison of human and GPT-4 evaluators. We report $ \mathrm{Acc}_p $ based on human and GPT-4 evaluation, and agreement between human and GPT-4 evaluation.}
\label{tab:results_agreement}
\end{table}

\subsection{Bias in Model Outputs}

Bias significantly influences model performance. We explore biases in model outputs on three dimensions: location \& orientation, timing and culture. We evaluate bias by calculating frequencies of three pairs of specific words, \textit{up} and \textit{down}, \textit{morning} and \textit{afternoon}, and \textit{appropriate} and \textit{not appropriate}, which are associated with the three dimensions, respectively. We evaluated the outputs of GPT-4V, Gemini, mPLUG-Owl-2, and LLaVA-1.5-13B, on a set of 88 queries whose groundtruth answers include these words. As shown in Table~\ref{tab:res_model_bias}, although frequencies of the two words in each pair are equal in groundtruth answers, all the MLLMs demonstrate biases in their responses. This indicates that models tend to rely on their biases rather than the provided context in a certain number of cases, contributing to their poor performance on \benchname benchmark, especially $ \mathrm{Acc}_p $ scores. Reducing bias is necessary for improving model capability on context-dependent visual comprehension. It also suggests that our design of data pairs and evaluation metric of $ \mathrm{Acc}_p $ effectively prevent models from scoring high based on their biases rather than a true understanding of contexts.

\subsection{Comparing Human and GPT-4 Evaluators}\label{sec:human_and_gpt4_eval}

We explore the possibility of utilizing GPT-4 as automatic evaluator. We compare the results of human and GPT-4 evaluation, as Table \ref{tab:results_agreement} shows. The first two columns show $ \mathrm{Acc}_p $ of models based on human and GPT-4 evaluation, and the third column shows agreement between human and GPT-4, which is calculated as follows:
$$
\mathrm{Agreement}=\frac{1}{n_q}\sum_{i=1}^{n_q}{(\mathrm{Eval_h}(O_i)=\mathrm{Eval_g}(O_i))}\text{,}
$$
where $ n_q $ is the total number of queries, $ \mathrm{Eval_h(O_i)} $ and $ \mathrm{Eval_g(O_i)} $ are evaluation results of human and GPT-4 on model output $ O_i $. From the table we can see that GPT-4 is consistent with human on evaluation of more than 90\% of queries. Overall, there is a high degree of consistency between human and GPT-4 evaluators in assessing model performance. GPT-4 is capable of conducting evaluation on \benchname benchmark instead of human. We report detailed results of GPT-4 evaluation in Appendix~\ref{sec:appendix_gpt4_eval}.

\section{Conclusion}

In this work, we introduce a new benchmark \benchname to evaluate capability of models to use context to enhance visual comprehension. Results suggest that MLLMs significantly fall behind human in context-dependent visual comprehension. Further analysis suggests that shortcomings in visual information extraction and bias in model outputs also account for low scores of MLLMs on \benchname. In summary, we pave a path for assessing capability of MLLMs on utilizing contextual information, which is a vital ability in real-world scenarios.

\section*{Limitations}

There are still some limitations that need to be addressed. First, contexts in this work are brief due to the limited ability of current MLLMs in effectively utilizing contextual information. We believe it is essential to assess capability of extracting information from lengthy and intricate context to enhance visual comprehension in the future, as MLLMs evolve and become more powerful. Second, despite covering a wide range of scenarios and five representative types of context, there are still other forms of contexts that we intend to explore in future endeavors. Moreover, delving into context from other modalities such as audio and depth presents an interesting avenue for exploration. Moving forward, we aim to establish an automated pipeline for generating queries with context information spanning a broader range of scenarios and modalities.

\section*{Acknowledgements}

This work is supported by the National Key R\&D Program of China (2022ZD0160502) and the National Natural Science Foundation of China (No. 61925601, 62276152). We thank Sijie Cheng, Zhicheng Guo for their advice in paper writing. We thank Zonghan Yang, Yile Wang and Yuanchi Zhang for their valuable discussions.

\bibliography{anthology, custom}

\begin{thebibliography}{54}
\expandafter\ifx\csname natexlab\endcsname\relax\def\natexlab#1{#1}\fi

\bibitem[{Antol et~al.(2015)Antol, Agrawal, Lu, Mitchell, Batra, Zitnick, and Parikh}]{antol2015vqa}
Stanislaw Antol, Aishwarya Agrawal, Jiasen Lu, Margaret Mitchell, Dhruv Batra, C~Lawrence Zitnick, and Devi Parikh. 2015.
\newblock {VQA}: Visual question answering.
\newblock In \emph{Proceedings of the IEEE International Conference on Computer Vision}, pages 2425--2433.

\bibitem[{Awadalla et~al.(2023)Awadalla, Gao, Gardner, Hessel, Hanafy, Zhu, Marathe, Bitton, Gadre, Sagawa et~al.}]{awadalla2023openflamingo}
Anas Awadalla, Irena Gao, Josh Gardner, Jack Hessel, Yusuf Hanafy, Wanrong Zhu, Kalyani Marathe, Yonatan Bitton, Samir Gadre, Shiori Sagawa, et~al. 2023.
\newblock {OpenFlamingo}: An open-source framework for training large autoregressive vision-language models.
\newblock \emph{arXiv preprint arXiv:2308.01390}.

\bibitem[{Bai et~al.(2023)Bai, Bai, Yang, Wang, Tan, Wang, Lin, Zhou, and Zhou}]{bai2023qwen}
Jinze Bai, Shuai Bai, Shusheng Yang, Shijie Wang, Sinan Tan, Peng Wang, Junyang Lin, Chang Zhou, and Jingren Zhou. 2023.
\newblock {Qwen-VL}: A frontier large vision-language model with versatile abilities.
\newblock \emph{arXiv preprint arXiv:2308.12966}.

\bibitem[{Baldassano et~al.(2016)Baldassano, Esteva, Fei-Fei, and Beck}]{baldassano2016two}
Christopher Baldassano, Andre Esteva, Li~Fei-Fei, and Diane~M Beck. 2016.
\newblock Two distinct scene-processing networks connecting vision and memory.
\newblock \emph{Eneuro}, 3(5).

\bibitem[{Bar and Aminoff(2003)}]{bar2003cortical}
Moshe Bar and Elissa Aminoff. 2003.
\newblock Cortical analysis of visual context.
\newblock \emph{Neuron}, 38(2):347--358.

\bibitem[{Chen et~al.(2023{\natexlab{a}})Chen, Li, Dong, Zhang, He, Wang, Zhao, and Lin}]{chen2023sharegpt4v}
Lin Chen, Jisong Li, Xiaoyi Dong, Pan Zhang, Conghui He, Jiaqi Wang, Feng Zhao, and Dahua Lin. 2023{\natexlab{a}}.
\newblock {ShareGPT4V}: Improving large multi-modal models with better captions.
\newblock \emph{arXiv preprint arXiv:2311.12793}.

\bibitem[{Chen et~al.(2023{\natexlab{b}})Chen, Sun, Song, and Luo}]{chen2023diffusiondet}
Shoufa Chen, Peize Sun, Yibing Song, and Ping Luo. 2023{\natexlab{b}}.
\newblock {DiffusionDet}: Diffusion model for object detection.
\newblock In \emph{Proceedings of the IEEE/CVF International Conference on Computer Vision}, pages 19830--19843.

\bibitem[{Chen et~al.(2015)Chen, Fang, Lin, Vedantam, Gupta, Doll{\'a}r, and Zitnick}]{chen2015microsoft}
Xinlei Chen, Hao Fang, Tsung-Yi Lin, Ramakrishna Vedantam, Saurabh Gupta, Piotr Doll{\'a}r, and C~Lawrence Zitnick. 2015.
\newblock {Microsoft COCO Captions}: Data collection and evaluation server.
\newblock \emph{arXiv preprint arXiv:1504.00325}.

\bibitem[{Chen et~al.(2023{\natexlab{c}})Chen, Zhou, Shen, Hong, Zhang, and Gan}]{chen2023see}
Zhenfang Chen, Qinhong Zhou, Yikang Shen, Yining Hong, Hao Zhang, and Chuang Gan. 2023{\natexlab{c}}.
\newblock See, think, confirm: Interactive prompting between vision and language models for knowledge-based visual reasoning.
\newblock \emph{arXiv preprint arXiv:2301.05226}.

\bibitem[{Cong et~al.(2023)Cong, Yang, and Rosenhahn}]{cong2023reltr}
Yuren Cong, Michael~Ying Yang, and Bodo Rosenhahn. 2023.
\newblock {RelTR}: Relation transformer for scene graph generation.
\newblock \emph{IEEE Transactions on Pattern Analysis and Machine Intelligence}.

\bibitem[{Cui et~al.(2023)Cui, Zhou, Yang, Wu, Zhang, Zou, and Yao}]{cui2023holistic}
Chenhang Cui, Yiyang Zhou, Xinyu Yang, Shirley Wu, Linjun Zhang, James Zou, and Huaxiu Yao. 2023.
\newblock Holistic analysis of hallucination in {GPT-4V(ision)}: Bias and interference challenges.
\newblock \emph{arXiv preprint arXiv:2311.03287}.

\bibitem[{Dai et~al.(2023)Dai, Li, Li, Tiong, Zhao, Wang, Li, Fung, and Hoi}]{dai2023instructblip}
Wenliang Dai, Junnan Li, Dongxu Li, Anthony Meng~Huat Tiong, Junqi Zhao, Weisheng Wang, Boyang Li, Pascale Fung, and Steven Hoi. 2023.
\newblock {InstructBLIP}: Towards general-purpose vision-language odels with instruction tuning.
\newblock \emph{arXiv preprint arXiv:2305.06500}.

\bibitem[{Das et~al.(2017)Das, Kottur, Gupta, Singh, Yadav, Moura, Parikh, and Batra}]{das2017visual}
Abhishek Das, Satwik Kottur, Khushi Gupta, Avi Singh, Deshraj Yadav, Jos{\'e}~MF Moura, Devi Parikh, and Dhruv Batra. 2017.
\newblock Visual dialog.
\newblock In \emph{Proceedings of the IEEE Conference on Computer Vision and Pattern Recognition}, pages 326--335.

\bibitem[{Denison et~al.(2018)Denison, Adler, Carrasco, and Ma}]{denison2018humans}
Rachel~N Denison, William~T Adler, Marisa Carrasco, and Wei~Ji Ma. 2018.
\newblock Humans incorporate attention-dependent uncertainty into perceptual decisions and confidence.
\newblock \emph{Proceedings of the National Academy of Sciences}, 115(43):11090--11095.

\bibitem[{Diwan et~al.(2023)Diwan, Anirudh, and Tembhurne}]{diwan2023object}
Tausif Diwan, G~Anirudh, and Jitendra~V Tembhurne. 2023.
\newblock Object detection using {YOLO}: Challenges, architectural successors, datasets and applications.
\newblock \emph{Multimedia Tools and Applications}, 82(6):9243--9275.

\bibitem[{Feng et~al.(2022)Feng, Sun, Xu, Zhao, Yang, Tao, Zhao, and Lin}]{feng2022mmdialog}
Jiazhan Feng, Qingfeng Sun, Can Xu, Pu~Zhao, Yaming Yang, Chongyang Tao, Dongyan Zhao, and Qingwei Lin. 2022.
\newblock {MMDialog}: A large-scale multi-turn dialogue dataset towards multi-modal open-domain conversation.
\newblock \emph{arXiv preprint arXiv:2211.05719}.

\bibitem[{Fu et~al.(2023{\natexlab{a}})Fu, Chen, Shen, Qin, Zhang, Lin, Yang, Zheng, Li, Sun et~al.}]{fu2023mme}
Chaoyou Fu, Peixian Chen, Yunhang Shen, Yulei Qin, Mengdan Zhang, Xu~Lin, Jinrui Yang, Xiawu Zheng, Ke~Li, Xing Sun, et~al. 2023{\natexlab{a}}.
\newblock {MME}: A comprehensive evaluation benchmark for multimodal large language models.
\newblock \emph{arXiv preprint arXiv:2306.13394}.

\bibitem[{Fu et~al.(2023{\natexlab{b}})Fu, Zhang, Lin, Wang, Gao, Luo, Huang, Zhang, Qiu, Ye et~al.}]{fu2023challenger}
Chaoyou Fu, Renrui Zhang, Haojia Lin, Zihan Wang, Timin Gao, Yongdong Luo, Yubo Huang, Zhengye Zhang, Longtian Qiu, Gaoxiang Ye, et~al. 2023{\natexlab{b}}.
\newblock A challenger to {GPT-4V}? early explorations of {Gemini} in visual expertise.
\newblock \emph{arXiv preprint arXiv:2312.12436}.

\bibitem[{Gemini~Team et~al.(2023)Gemini~Team, Borgeaud, Wu, Alayrac, Yu, Soricut, Schalkwyk, Dai, Hauth et~al.}]{gemini2023gemini}
G~Gemini~Team, Sebastian Borgeaud, Yonghui Wu, Jean-Baptiste Alayrac, Jiahui Yu, Radu Soricut, Johan Schalkwyk, Andrew~M Dai, Anja Hauth, et~al. 2023.
\newblock {Gemini}: A family of highly capable multimodal models.
\newblock \emph{arXiv preprint arXiv:2312.11805}.

\bibitem[{Goh et~al.(2004)Goh, Siong, Park, Gutchess, Hebrank, and Chee}]{goh2004cortical}
Joshua~OS Goh, Soon~Chun Siong, Denise Park, Angela Gutchess, Andy Hebrank, and Michael~WL Chee. 2004.
\newblock Cortical areas involved in object, background, and object-background processing revealed with functional magnetic resonance adaptation.
\newblock \emph{Journal of Neuroscience}, 24(45):10223--10228.

\bibitem[{Guan et~al.(2023)Guan, Liu, Li, Wang, Yacoob, and Zhou}]{guan2023hallusionbench}
Tianrui Guan, Fuxiao Liu, Xiyang Wu Ruiqi Xian~Zongxia Li, Xiaoyu Liu~Xijun Wang, Lichang Chen Furong Huang~Yaser Yacoob, and Dinesh Manocha~Tianyi Zhou. 2023.
\newblock {HallusionBench}: An advanced diagnostic suite for entangled language hallucination \& visual illusion in large vision-language models.
\newblock \emph{arXiv e-prints}, pages arXiv--2310.

\bibitem[{Gupta and Kembhavi(2023)}]{gupta2023visual}
Tanmay Gupta and Aniruddha Kembhavi. 2023.
\newblock {Visual Programming}: Compositional visual reasoning without training.
\newblock In \emph{Proceedings of the IEEE/CVF Conference on Computer Vision and Pattern Recognition}, pages 14953--14962.

\bibitem[{Krishna et~al.(2017)Krishna, Zhu, Groth, Johnson, Hata, Kravitz, Chen, Kalantidis, Li, Shamma et~al.}]{krishna2017visual}
Ranjay Krishna, Yuke Zhu, Oliver Groth, Justin Johnson, Kenji Hata, Joshua Kravitz, Stephanie Chen, Yannis Kalantidis, Li-Jia Li, David~A Shamma, et~al. 2017.
\newblock {Visual Genome}: Connecting language and vision using crowdsourced dense image annotations.
\newblock \emph{International journal of computer vision}, 123:32--73.

\bibitem[{Li et~al.(2023{\natexlab{a}})Li, Zhang, Chen, Wang, Pu, Yang, Li, and Liu}]{li2023mimic}
Bo~Li, Yuanhan Zhang, Liangyu Chen, Jinghao Wang, Fanyi Pu, Jingkang Yang, Chunyuan Li, and Ziwei Liu. 2023{\natexlab{a}}.
\newblock {MIMIC-IT}: Multi-modal in-context instruction tuning.
\newblock \emph{arXiv preprint arXiv:2306.05425}.

\bibitem[{Li et~al.(2023{\natexlab{b}})Li, Wang, Wang, Ge, Ge, and Shan}]{li2023seed}
Bohao Li, Rui Wang, Guangzhi Wang, Yuying Ge, Yixiao Ge, and Ying Shan. 2023{\natexlab{b}}.
\newblock {SEED-Bench}: Benchmarking multimodal llms with generative comprehension.
\newblock \emph{arXiv preprint arXiv:2307.16125}.

\bibitem[{Li et~al.(2023{\natexlab{c}})Li, Pan, Ge, Gao, Zhang, Ji, Zhang, Chua, Tang, and Zhuang}]{li2023fine}
Juncheng Li, Kaihang Pan, Zhiqi Ge, Minghe Gao, Hanwang Zhang, Wei Ji, Wenqiao Zhang, Tat-Seng Chua, Siliang Tang, and Yueting Zhuang. 2023{\natexlab{c}}.
\newblock Fine-tuning multimodal llms to follow zeroshot demonstrative instructions.
\newblock \emph{arXiv preprint arXiv:2308.04152}, 3.

\bibitem[{Li et~al.(2023{\natexlab{d}})Li, Li, Savarese, and Hoi}]{li2023blip}
Junnan Li, Dongxu Li, Silvio Savarese, and Steven Hoi. 2023{\natexlab{d}}.
\newblock {BLIP-2}: Bootstrapping language-image pre-training with frozen image encoders and large language models.
\newblock \emph{arXiv preprint arXiv:2301.12597}.

\bibitem[{Liu et~al.(2023{\natexlab{a}})Liu, Li, Li, and Lee}]{liu2023improved}
Haotian Liu, Chunyuan Li, Yuheng Li, and Yong~Jae Lee. 2023{\natexlab{a}}.
\newblock Improved baselines with visual instruction tuning.
\newblock \emph{arXiv preprint arXiv:2310.03744}.

\bibitem[{Liu et~al.(2023{\natexlab{b}})Liu, Li, Wu, and Lee}]{liu2023visual}
Haotian Liu, Chunyuan Li, Qingyang Wu, and Yong~Jae Lee. 2023{\natexlab{b}}.
\newblock Visual instruction tuning.
\newblock \emph{arXiv preprint arXiv:2304.08485}.

\bibitem[{Liu et~al.(2023{\natexlab{c}})Liu, Li, and Lin}]{liu2023cross}
Yang Liu, Guanbin Li, and Liang Lin. 2023{\natexlab{c}}.
\newblock Cross-modal causal relational reasoning for event-level visual question answering.
\newblock \emph{IEEE Transactions on Pattern Analysis and Machine Intelligence}.

\bibitem[{Liu et~al.(2023{\natexlab{d}})Liu, Duan, Zhang, Li, Zhang, Zhao, Yuan, Wang, He, Liu et~al.}]{liu2023mmbench}
Yuan Liu, Haodong Duan, Yuanhan Zhang, Bo~Li, Songyang Zhang, Wangbo Zhao, Yike Yuan, Jiaqi Wang, Conghui He, Ziwei Liu, et~al. 2023{\natexlab{d}}.
\newblock {MMBench}: Is your multi-modal model an all-around player?
\newblock \emph{arXiv preprint arXiv:2307.06281}.

\bibitem[{Luo et~al.(2023)Luo, Li, Pan, Yao, Feng, Chao, and Mei}]{luo2023semantic}
Jianjie Luo, Yehao Li, Yingwei Pan, Ting Yao, Jianlin Feng, Hongyang Chao, and Tao Mei. 2023.
\newblock Semantic-conditional diffusion networks for image captioning.
\newblock In \emph{Proceedings of the IEEE/CVF Conference on Computer Vision and Pattern Recognition}, pages 23359--23368.

\bibitem[{OpenAI(2023)}]{openai2023gpt4vision}
OpenAI. 2023.
\newblock \href {https://cdn.openai.com/papers/GPTV_System_Card.pdf} {{GPT-4V(ision) System Card}}.

\bibitem[{Purushwalkam et~al.(2021)Purushwalkam, Gari, Ithapu, Schissler, Robinson, Gupta, and Grauman}]{purushwalkam2021audio}
Senthil Purushwalkam, Sebastia Vicenc~Amengual Gari, Vamsi~Krishna Ithapu, Carl Schissler, Philip Robinson, Abhinav Gupta, and Kristen Grauman. 2021.
\newblock Audio-visual floorplan reconstruction.
\newblock In \emph{Proceedings of the IEEE/CVF International Conference on Computer Vision}, pages 1183--1192.

\bibitem[{Rajeswar et~al.(2022)Rajeswar, Rodriguez, Singhal, Vazquez, and Courville}]{rajeswar2022multi}
Sai Rajeswar, Pau Rodriguez, Soumye Singhal, David Vazquez, and Aaron Courville. 2022.
\newblock Multi-label iterated learning for image classification with label ambiguity.
\newblock In \emph{Proceedings of the IEEE/CVF Conference on Computer Vision and Pattern Recognition}, pages 4783--4793.

\bibitem[{Sanders et~al.(2022)Sanders, Kriz, Liu, and Van~Durme}]{sanders2022ambiguous}
Kate Sanders, Reno Kriz, Anqi Liu, and Benjamin Van~Durme. 2022.
\newblock Ambiguous images with human judgments for robust visual event classification.
\newblock \emph{Advances in Neural Information Processing Systems}, 35:2637--2650.

\bibitem[{Sargsyan et~al.(2023)Sargsyan, Navasardyan, Xu, and Shi}]{sargsyan2023mi}
Andranik Sargsyan, Shant Navasardyan, Xingqian Xu, and Humphrey Shi. 2023.
\newblock {MI-GAN}: A simple baseline for image inpainting on mobile devices.
\newblock In \emph{Proceedings of the IEEE/CVF International Conference on Computer Vision}, pages 7335--7345.

\bibitem[{Schmarje et~al.(2022)Schmarje, Grossmann, Zelenka, Dippel, Kiko, Oszust, Pastell, Stracke, Valros, Volkmann et~al.}]{schmarje2022one}
Lars Schmarje, Vasco Grossmann, Claudius Zelenka, Sabine Dippel, Rainer Kiko, Mariusz Oszust, Matti Pastell, Jenny Stracke, Anna Valros, Nina Volkmann, et~al. 2022.
\newblock Is one annotation enough?-a data-centric image classification benchmark for noisy and ambiguous label estimation.
\newblock \emph{Advances in Neural Information Processing Systems}, 35:33215--33232.

\bibitem[{Shao et~al.(2023)Shao, Yu, Wang, and Yu}]{shao2023prompting}
Zhenwei Shao, Zhou Yu, Meng Wang, and Jun Yu. 2023.
\newblock Prompting large language models with answer heuristics for knowledge-based visual question answering.
\newblock In \emph{Proceedings of the IEEE/CVF Conference on Computer Vision and Pattern Recognition}, pages 14974--14983.

\bibitem[{Sun et~al.(2023)Sun, Cui, Zhang, Zhang, Yu, Luo, Wang, Rao, Liu, Huang et~al.}]{sun2023generative}
Quan Sun, Yufeng Cui, Xiaosong Zhang, Fan Zhang, Qiying Yu, Zhengxiong Luo, Yueze Wang, Yongming Rao, Jingjing Liu, Tiejun Huang, et~al. 2023.
\newblock Generative multimodal models are in-context learners.
\newblock \emph{arXiv preprint arXiv:2312.13286}.

\bibitem[{Vo(2021)}]{vo2021meaning}
Melissa Le-Hoa Vo. 2021.
\newblock The meaning and structure of scenes.
\newblock \emph{Vision Research}, 181:10--20.

\bibitem[{Wang et~al.(2023)Wang, Xie, Wu, Jia, and Li}]{wang2023controllable}
Ning Wang, Jiahao Xie, Jihao Wu, Mingbo Jia, and Linlin Li. 2023.
\newblock Controllable image captioning via prompting.
\newblock In \emph{Proceedings of the AAAI Conference on Artificial Intelligence}, pages 2617--2625.

\bibitem[{Wang and Zhu(2023)}]{wang2023context}
Xuan Wang and Zhigang Zhu. 2023.
\newblock Context understanding in computer vision: A survey.
\newblock \emph{Computer Vision and Image Understanding}, 229:103646.

\bibitem[{Xiang et~al.(2023)Xiang, Zou, Nawaz, Huang, Zhang, and Yu}]{xiang2023deep}
Hanyu Xiang, Qin Zou, Muhammad~Ali Nawaz, Xianfeng Huang, Fan Zhang, and Hongkai Yu. 2023.
\newblock Deep learning for image inpainting: A survey.
\newblock \emph{Pattern Recognition}, 134:109046.

\bibitem[{Xiao et~al.(2023)Xiao, Guo, Zhou, Zhao, Yu, Chen, and Wang}]{xiao2023tiny}
Jinsheng Xiao, Haowen Guo, Jian Zhou, Tao Zhao, Qiuze Yu, Yunhua Chen, and Zhongyuan Wang. 2023.
\newblock Tiny object detection with context enhancement and feature purification.
\newblock \emph{Expert Systems with Applications}, 211:118665.

\bibitem[{Yang et~al.(2018)Yang, Lu, Lee, Batra, and Parikh}]{yang2018graph}
Jianwei Yang, Jiasen Lu, Stefan Lee, Dhruv Batra, and Devi Parikh. 2018.
\newblock Graph r-cnn for scene graph generation.
\newblock In \emph{Proceedings of the European Conference on Computer Vision (ECCV)}.

\bibitem[{Ye et~al.(2023)Ye, Xu, Ye, Yan, Liu, Qian, Zhang, Huang, and Zhou}]{ye2023mplugowl2}
Qinghao Ye, Haiyang Xu, Jiabo Ye, Ming Yan, Haowei Liu, Qi~Qian, Ji~Zhang, Fei Huang, and Jingren Zhou. 2023.
\newblock {mPLUG-Owl2}: Revolutionizing multi-modal large language model with modality collaboration.
\newblock \emph{arXiv preprint arXiv:2311.04257}.

\bibitem[{Zellers et~al.(2019)Zellers, Bisk, Farhadi, and Choi}]{zellers2019recognition}
Rowan Zellers, Yonatan Bisk, Ali Farhadi, and Yejin Choi. 2019.
\newblock From recognition to cognition: Visual commonsense reasoning.
\newblock In \emph{Proceedings of the IEEE/CVF Conference on Computer Vision and Pattern Recognition}, pages 6720--6731.

\bibitem[{Zhang et~al.(2023)Zhang, Zhai, Li, Zhou, and Lin}]{zhang2023image}
Xiaobo Zhang, Donghai Zhai, Tianrui Li, Yuxin Zhou, and Yang Lin. 2023.
\newblock Image inpainting based on deep learning: A review.
\newblock \emph{Information Fusion}, 90:74--94.

\bibitem[{Zhao et~al.(2023)Zhao, Cai, Si, Ma, An, Chen, Liu, Wang, Han, and Chang}]{zhao2023mmicl}
Haozhe Zhao, Zefan Cai, Shuzheng Si, Xiaojian Ma, Kaikai An, Liang Chen, Zixuan Liu, Sheng Wang, Wenjuan Han, and Baobao Chang. 2023.
\newblock {MMICL}: Empowering vision-language model with multi-modal in-context learning.
\newblock \emph{arXiv preprint arXiv:2309.07915}.

\bibitem[{Zheng et~al.(2023)Zheng, Lyu, Gao, Dai, and Song}]{zheng2023prototype}
Chaofan Zheng, Xinyu Lyu, Lianli Gao, Bo~Dai, and Jingkuan Song. 2023.
\newblock Prototype-based embedding network for scene graph generation.
\newblock In \emph{Proceedings of the IEEE/CVF Conference on Computer Vision and Pattern Recognition}, pages 22783--22792.

\bibitem[{Zhu et~al.(2023)Zhu, Chen, Shen, Li, and Elhoseiny}]{zhu2023minigpt}
Deyao Zhu, Jun Chen, Xiaoqian Shen, Xiang Li, and Mohamed Elhoseiny. 2023.
\newblock {MiniGPT-4}: Enhancing vision-language understanding with advanced large language models.
\newblock \emph{arXiv preprint arXiv:2304.10592}.

\bibitem[{Zhu et~al.(2022)Zhu, Zhang, Jiang, Dang, Hou, Shen, Feng, Zhao, Miao, Shah et~al.}]{zhu2022scene}
Guangming Zhu, Liang Zhang, Youliang Jiang, Yixuan Dang, Haoran Hou, Peiyi Shen, Mingtao Feng, Xia Zhao, Qiguang Miao, Syed Afaq~Ali Shah, et~al. 2022.
\newblock Scene graph generation: A comprehensive survey.
\newblock \emph{arXiv preprint arXiv:2201.00443}.

\bibitem[{Zou et~al.(2023)Zou, Chen, Shi, Guo, and Ye}]{zou2023object}
Zhengxia Zou, Keyan Chen, Zhenwei Shi, Yuhong Guo, and Jieping Ye. 2023.
\newblock Object detection in 20 years: A survey.
\newblock \emph{Proceedings of the IEEE}.

\end{thebibliography}

\clearpage

\appendix

\section{Consistency of Human Evaluation}\label{sec:appendix_consistency_human_eval}

A total of four individuals participated in human evaluation of model outputs. We examine the consistency of their assessments. A random sample of 100 queries out of 706 was selected for evaluation. Each of the four individuals made judgments on the model outputs of these 100 queries individually. A groundtruth value of 1 was assigned when the majority agreed the answer was correct, and a value of 0 was assigned when the majority agreed the answer was incorrect. In cases of a tie, a fifth individual was introduced. The average consistency between human judgments and groundtruths was then calculated as follows:
$$
\mathrm{Consistency}=\frac{1}{4n_q}\sum_{i=1}^{n_q}\sum_{j=1}^{4}{(\mathrm{Eval}_j(O_i)=\mathrm{GT}_i)}\text{,}
$$
where $ n_q $ represents the total number of queries, $ \mathrm{Eval}_j(O_i) $ denotes the evaluation of individual $ j $ on model output of query $ i $ and $ 
\mathrm{GT}_i $ indicates the groundtruth assessment of model output of query $ i $. The results are shown in Table~\ref{tab:results_consistency_human}.

\begin{table}[ht]
\centering
\small
\begin{tabular}{lcc}
\toprule
\textbf{Model} & \textbf{Human Consistency} \\\midrule
GPT-4V & 98.3 \\
Gemini & 99.3 \\
mPLUG-Owl-2 & 97.3 \\
MiniGPT4-7B & 95.3 \\
LLaVA-1.5-7B & 97.5 \\
InstructBLIP-7B & 99.5 \\
Otter-7B & 97.0 \\
LLaVA-7B & 96.8 \\
Qwen-VL-Chat & 96.5 \\
OpenFlamingo-7B & 99.0 \\
BLIP-2-6.7B & 99.5 \\
LLaVA-1.5-13B & 98.8 \\
BLIP-2-11B & 97.0 \\
InstructBLIP-13B & 98.8 \\\bottomrule
\end{tabular}
\caption{Consistency of human evaluation on \benchname.}
\label{tab:results_consistency_human}
\end{table}

The results indicate that the consistency of human evaluation is high. As a result, to enhance efficiency in the remaining manual evaluation tasks in this study, we distributed tasks evenly among the four individuals instead of employing the majority vote method. Moreover, agreement between human and GPT-4 evaluators , as portrayed in Table~\ref{tab:results_agreement}, is close to human consistency, indicating proficiency of GPT-4 on evaluating model outputs is comparable to that of humans.

\section{Human Scores on \benchname}\label{sec:appendix_human_test}

We enlisted participation of five volunteers to respond to the questions in \benchname. Their average age was 31 years old and they possess proficient daily English communication skills. As they are volunteers, they are not paid for the test. Before the test, all the volunteers were completely unaware of the questions and answers.

\begin{table}[ht]
\centering
\small
\begin{tabular}{lcc}
\toprule
\textbf{Testee} & \textbf{Acc\textsubscript{\textit{p}}} & \textbf{Acc\textsubscript{\textit{q}}} \\\midrule
Human 1 & 91.8 & 92.6 \\
Human 2 & 82.7 & 84.4 \\
Human 3 & 89.0 & 89.4 \\
Human 4 & 83.6 & 84.6 \\
Human 5 & 83.9 & 87.4 \\
Average & 86.2 & 87.7 \\\bottomrule
\end{tabular}
\caption{Human performance on \benchname.}
\label{tab:results_human}
\end{table}

As shown in Table~\ref{tab:results_human}, humans perform well in \benchname benchmark, and achieve significantly higher scores than MLLMs. However, they do not reach 100\% scores. Upon examining the results, we identify two primary reasons for this. Firstly, some questions require a lengthy chain of reasoning, which poses challenges for humans as well. Secondly, humans sometimes lack specific background knowledge, particularly in questions related to cultural backgrounds.

\section{Cases of Model Outputs When Removing Context}\label{sec:appendix_case_remove_context}

\begin{figure*}[ht]
    \begin{minipage}[t]{0.33\textwidth}
        \centering
        \includegraphics[width=1\textwidth]{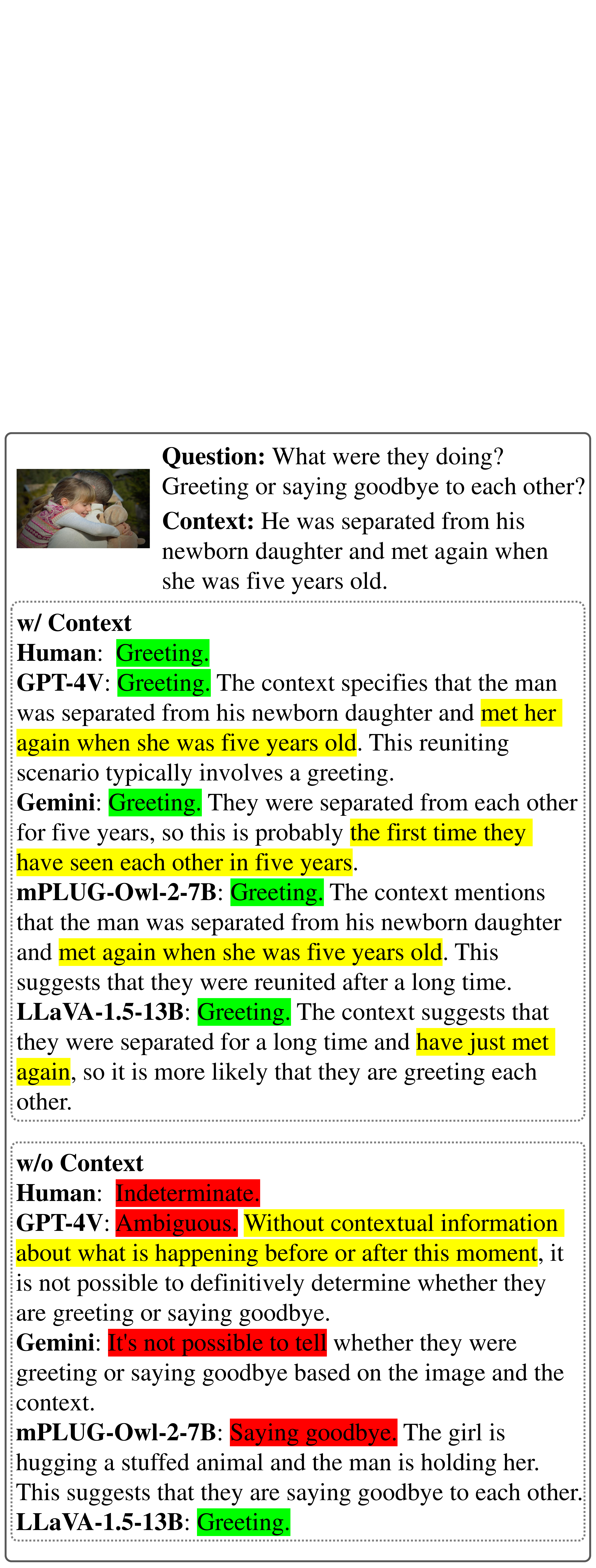}
    \end{minipage}
    \begin{minipage}[t]{0.33\textwidth}
        \centering
        \includegraphics[width=1\textwidth]{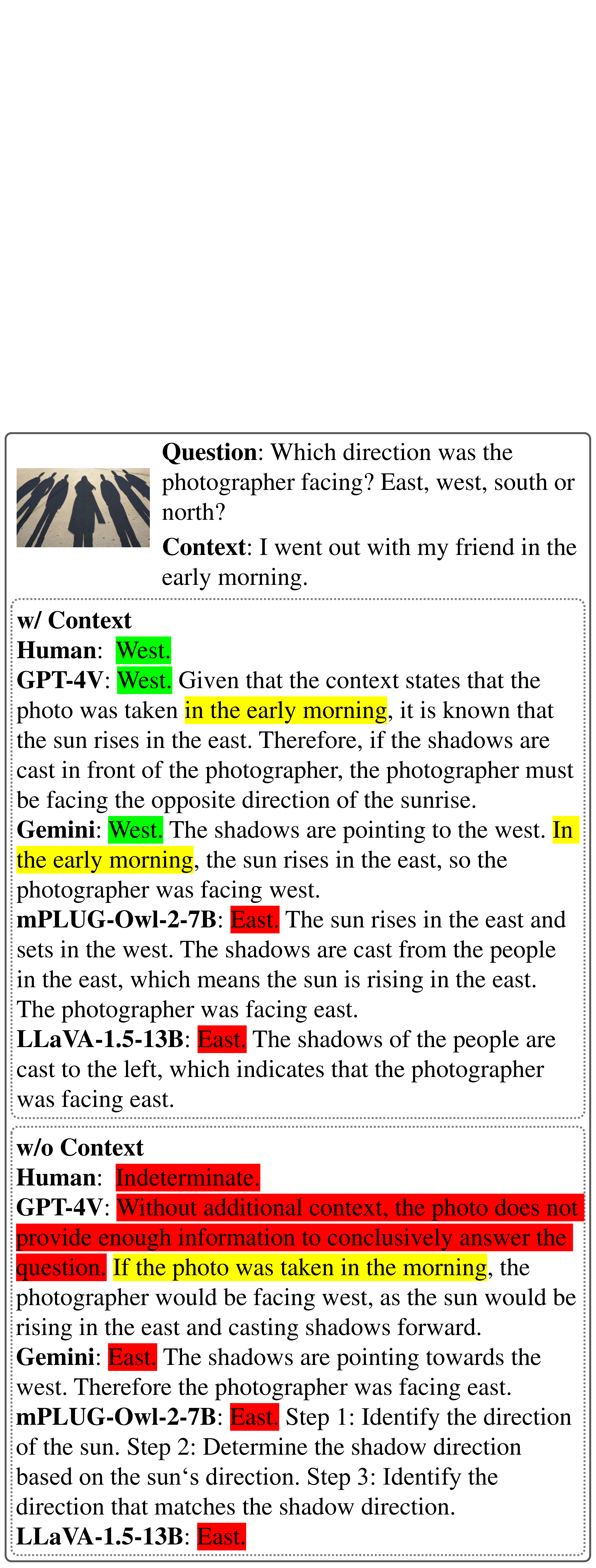}
    \end{minipage}
    \begin{minipage}[t]{0.33\textwidth}
        \centering
        \includegraphics[width=1\textwidth]{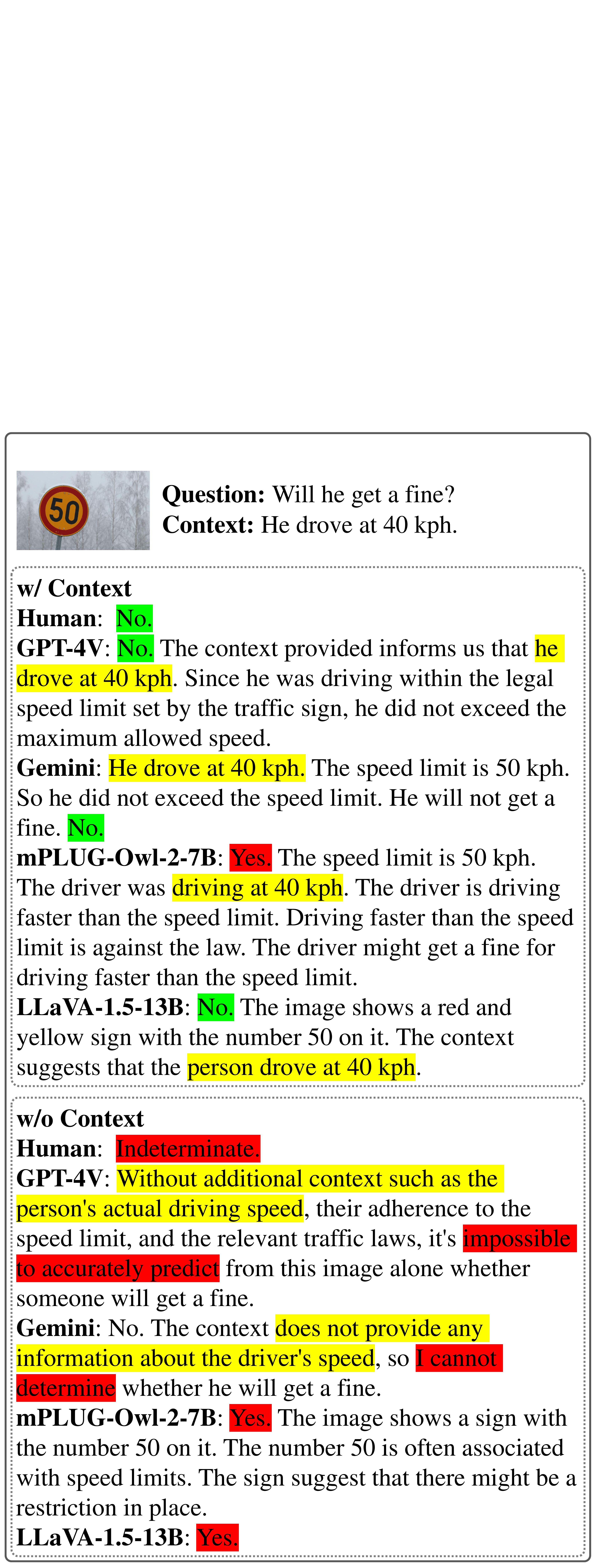}
    \end{minipage}
    \caption{Three cases to show alterations in model outputs resulting from the removal of context information.}
    \label{fig:remove_context}
\end{figure*}

We provide three cases in Figure~\ref{fig:remove_context} to show alterations in model outputs resulting from the removal of context information. We use color \colorbox{green}{green} and \colorbox{red}{red} to emphasize model outputs in agreement and disagreement with the groundtruth answers, and use \colorbox{yellow}{yellow} to highlight the words associated with context information in model outputs.

\section{Cases of Model Outputs When Replacing Images with Captions}\label{sec:appendix_case_replace_image_with_caption}

\begin{figure*}[ht]
    \begin{minipage}[t]{0.33\textwidth}
        \centering
        \includegraphics[width=1\textwidth]{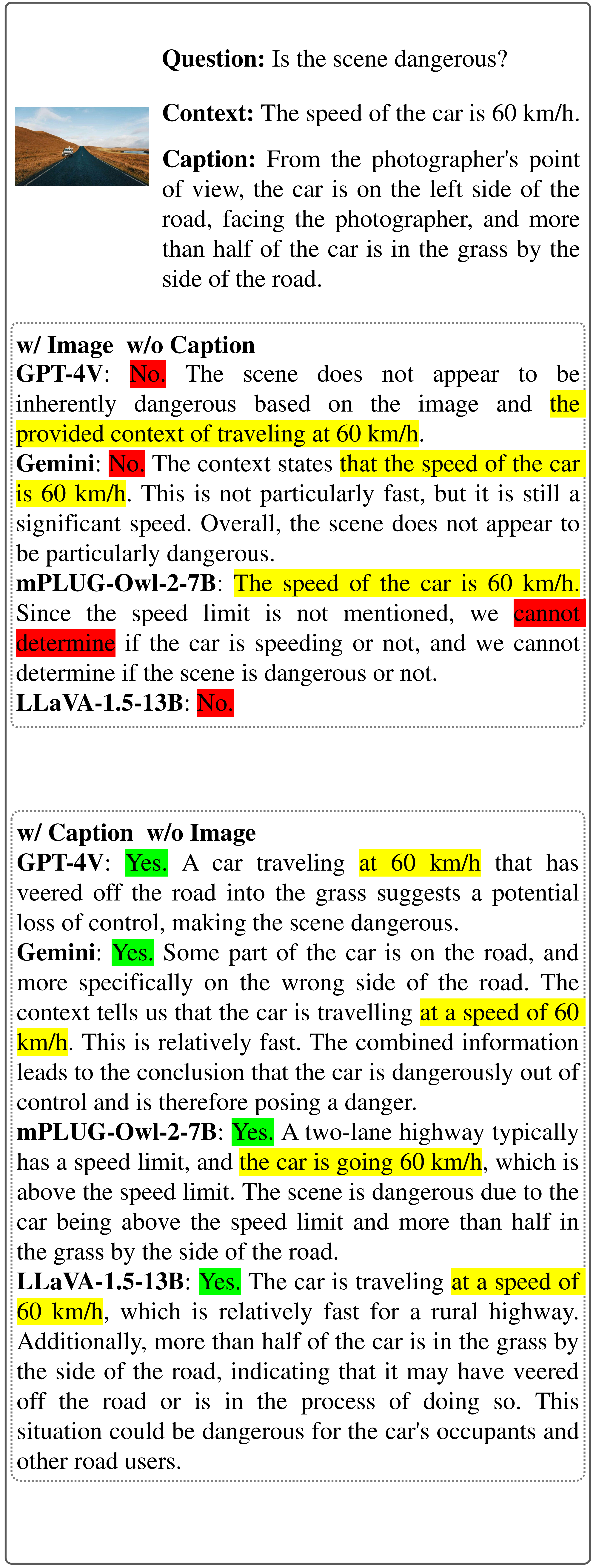}
    \end{minipage}
    \begin{minipage}[t]{0.33\textwidth}
        \centering
        \includegraphics[width=1\textwidth]{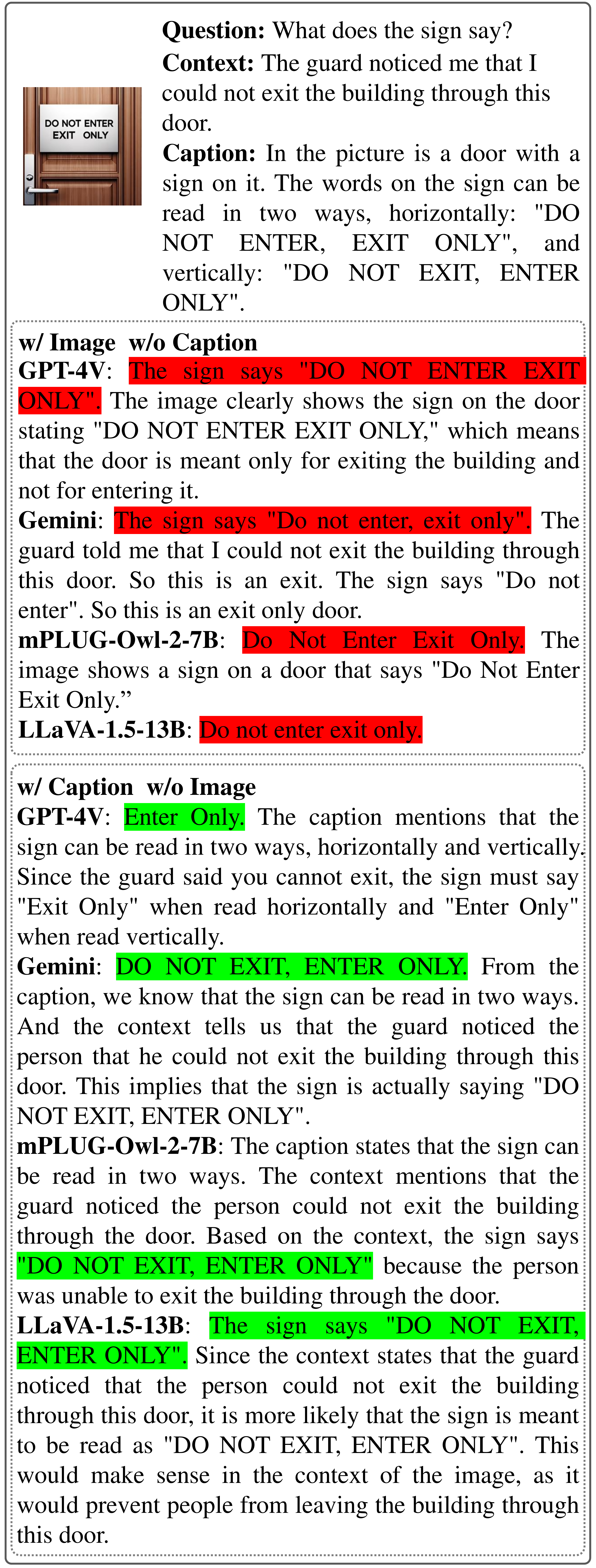}
    \end{minipage}
    \begin{minipage}[t]{0.33\textwidth}
        \centering
        \includegraphics[width=1\textwidth]{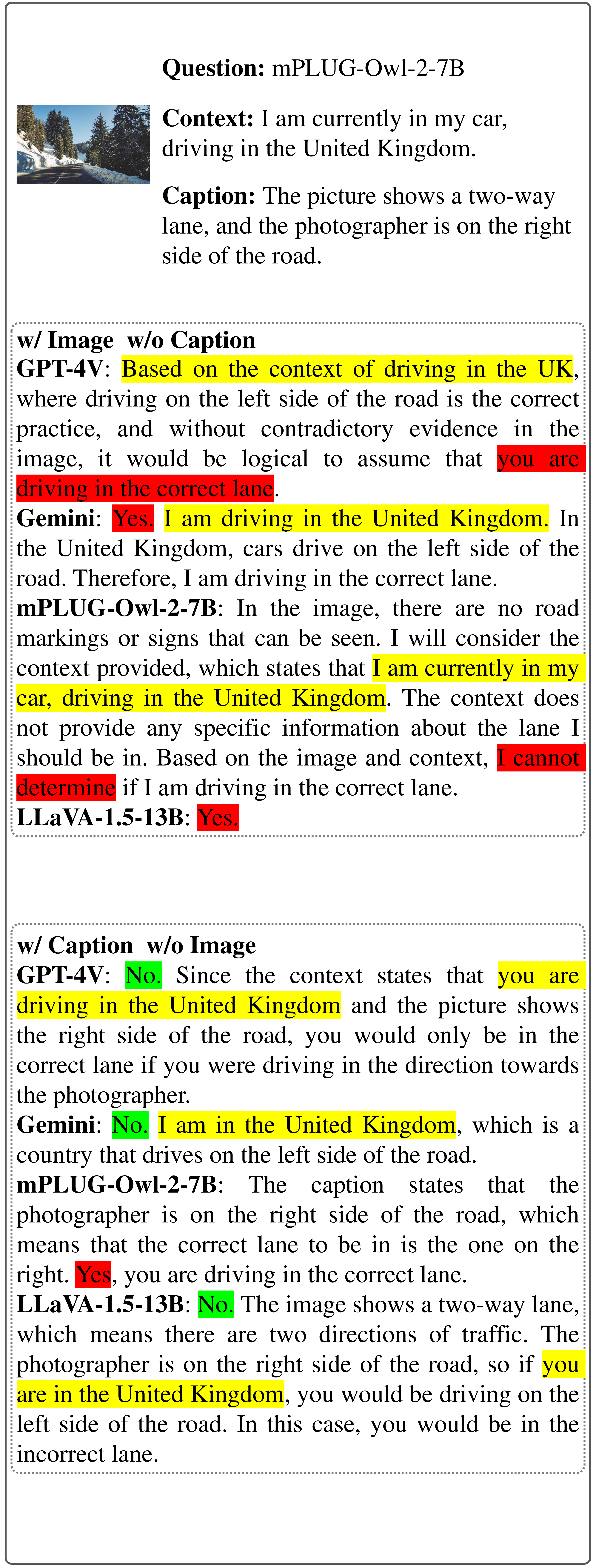}
    \end{minipage}
    \caption{Three cases to show alterations in model outputs when replacing images with captions.}
    \label{fig:replace_caption}
\end{figure*}

We provide three cases in Figure~\ref{fig:replace_caption} to show alterations in model outputs when replacing images with captions. We employ the identical highlighting markers as those utilized in Appendix~\ref{sec:appendix_case_remove_context}.

\section{Detailed Results of GPT-4 Evaluation}\label{sec:appendix_gpt4_eval}

We report detailed results based on GPT-4 evaluation in Table~\ref{tab:results_gpt4}. We report $ \mathrm{Acc}_p $ and $ \mathrm{Acc}_q $. The results exhibit a strong alignment with the results in Table~\ref{tab:results} based on human evaluation.

\begin{table*}[ht]
\centering
\small
\begin{tabular}{l|cccccccccc|cc}
\toprule
\multirow{2}{*}{\textbf{Model}} & \multicolumn{2}{c}{\textbf{Loc \& Ori}} & \multicolumn{2}{c}{\textbf{Temporal}} & \multicolumn{2}{c}{\textbf{Cultural}} & \multicolumn{2}{c}{\textbf{Attributes}} & \multicolumn{2}{c|}{\textbf{Relationships}} & \multicolumn{2}{c}{\textbf{Overall}} \\
& \textbf{Acc\textsubscript{\textit{p}}} & \textbf{Acc\textsubscript{\textit{q}}} & \textbf{Acc\textsubscript{\textit{p}}} & \textbf{Acc\textsubscript{\textit{q}}} & \textbf{Acc\textsubscript{\textit{p}}} & \textbf{Acc\textsubscript{\textit{q}}} & \textbf{Acc\textsubscript{\textit{p}}} & \textbf{Acc\textsubscript{\textit{q}}} & \textbf{Acc\textsubscript{\textit{p}}} & \textbf{Acc\textsubscript{\textit{q}}} & \textbf{Acc\textsubscript{\textit{p}}} & \textbf{Acc\textsubscript{\textit{q}}} \\\midrule\noalign{\vskip -3pt}
\rowcolor{gray!20}\multicolumn{13}{c}{\textbf{API-based Models}} \\\noalign{\vskip -2pt}\midrule
GPT-4V & 33.3 & 53.6 & 28.4 & 50.5 & 21.3 & 53.2 & 25.3 & 54.0 & 50.0 & 69.2 & 31.2 & 55.1 \\
Gemini & 20.2 & 48.8 & 27.4 & 50.0 & 21.3 & 54.3 & 22.7 & 51.3 & 30.8 & 54.8 & 24.4 & 51.3  \\\midrule\noalign{\vskip -3pt}
\rowcolor{gray!20}\multicolumn{13}{c}{\textbf{Open-source \textasciitilde{}13B Models}} \\\noalign{\vskip -2pt}\midrule
LLaVA-1.5-13B & \hphantom{0}6.0 & 41.1 & \hphantom{0}3.2 & 43.2 & 12.8 & 46.8 & 13.3 & 50.0 & 11.5 & 53.8 & \hphantom{0}8.5 & 46.2 \\
BLIP-2-11B & \hphantom{0}6.0 & 34.5 & 10.5 & 44.2 & \hphantom{0}4.3 & 30.9 & \hphantom{0}6.7 & 40.7 & 11.5 & 47.1 & \hphantom{0}8.0 & 39.8 \\
InstructBLIP-13B & \hphantom{0}6.0 & 39.9 & \hphantom{0}2.1 & 41.1 & \hphantom{0}6.4 & 46.8 & \hphantom{0}4.0 & 44.7 & \hphantom{0}5.8 & 48.1 & \hphantom{0}4.5 & 43.3 \\\midrule\noalign{\vskip -3pt}
\rowcolor{gray!20}\multicolumn{13}{c}{\textbf{Open-source \textasciitilde{}7B Models}} \\\noalign{\vskip -2pt}\midrule
mPLUG-Owl-2 & 13.1 & 39.9 & \hphantom{0}9.5 & 40.0 & \hphantom{0}4.3 & 41.5 & \hphantom{0}9.3 & 42.7 & 11.5 & 48.1 & \hphantom{0}9.9 & 41.9 \\
MiniGPT4-7B & 10.7 & 34.5 & \hphantom{0}4.2 & 32.1 & \hphantom{0}0.0 & 27.7 & \hphantom{0}8.0 & 32.7 & \hphantom{0}9.6 & 43.3 & \hphantom{0}6.8 & 33.9 \\
LLaVA-1.5-7B & \hphantom{0}8.3 & 37.5 & \hphantom{0}1.1 & 36.3 & \hphantom{0}2.1 & 40.4 & \hphantom{0}9.3 & 37.3 & \hphantom{0}7.7 & 48.1 & \hphantom{0}5.7 & 39.1 \\
InstructBLIP-7B & \hphantom{0}1.2 & 34.5 & \hphantom{0}5.3 & 43.7 & \hphantom{0}0.0 & 45.7 & \hphantom{0}4.0 & 44.0 & 11.5 & 47.1 & \hphantom{0}4.2 & 42.4 \\
Otter-7B & \hphantom{0}2.4 & 31.5 & \hphantom{0}3.2 & 35.3 & \hphantom{0}0.0 & 23.4 & \hphantom{0}1.3 & 27.3 & \hphantom{0}5.8 & 34.6 & \hphantom{0}2.5 & 31.0 \\
LLaVA-7B & \hphantom{0}2.4 & 29.8 & \hphantom{0}4.2 & 33.7 & \hphantom{0}0.0 & 17.0 & \hphantom{0}2.7 & 33.3 & \hphantom{0}1.9 & 37.5 & \hphantom{0}2.5 & 31.0 \\
Qwen-VL-Chat & \hphantom{0}4.8 & 23.8 & \hphantom{0}3.2 & 23.7 & \hphantom{0}0.0 & 23.4 & \hphantom{0}1.3 & 32.0 & \hphantom{0}7.7 & 33.7 & \hphantom{0}3.4 & 26.9 \\
OpenFlamingo-7B & \hphantom{0}2.4 & 40.5 & \hphantom{0}2.1 & 38.9 & \hphantom{0}0.0 & 27.7 & \hphantom{0}5.3 & 36.0 & \hphantom{0}5.8 & 47.1 & \hphantom{0}3.1 & 38.4 \\
BLIP-2-6.7B & \hphantom{0}0.0 & 42.3 & \hphantom{0}1.1 & 43.2 & \hphantom{0}4.3 & 48.9 & \hphantom{0}4.0 & 46.7 & \hphantom{0}5.8 & 51.0 & \hphantom{0}2.5 & 45.6 \\\bottomrule
\end{tabular}
\caption{Results of MLLMs on \benchname. All the model outputs are assessed by GPT-4.}
\label{tab:results_gpt4}
\end{table*}

\begin{table*}[ht]
\centering
\small
\begin{tabularx}{\linewidth}{m{1cm}<{\centering}m{\dimexpr\textwidth-1cm-4\tabcolsep}}
\toprule
\multicolumn{2}{c}{\textbf{Prompt for Model Inference}} \\\midrule
\makecell{w/ DI \\ w/ CoT} &
I'll give you an image and some additional context, which provides information closely related to the scene of the picture. Please answer my question based on the image and the context. Be sure to refer to the context and extract necessary information from it to help you answer the question because it contains helpful information that is not included in the image.

Your answer should contain two parts. Two parts should be separated by a newline. In the first part, please think of the question step by step based on the image and context and output your reasoning process. In the second part, please summarize your reasoning process and directly answer the question in a single word or phrase.

Context: [CONTEXT HERE]

Question: [QUESTION HERE] \\\midrule
\makecell{w/ DI \\ w/o CoT} &
I'll give you an image and some additional context, which provides information closely related to the scene of the picture. Please answer my question based on the image and the context. Be sure to refer to the context and extract necessary information from it to help you answer the question because it contains helpful information that is not included in the image.

Please answer in a single word or phrase.

Context: [CONTEXT HERE]

Question: [QUESTION HERE] \\\midrule
\makecell{w/o DI \\ w/ CoT} &
I'll give you an image and some additional context, which provides information closely related to the scene of the picture. Please answer my question based on the image and the context.

Your answer should contain two parts. Two parts should be seperated by a newline. In the first part, please think of the question step by step based on the image and context and output your reasoning process. In the second part, please summarize your reasoning process and directly answer the question in a single word or phrase.

Context: [CONTEXT HERE]

Question: [QUESTION HERE] \\\midrule
\makecell{w/o DI \\ w/o CoT} &
I'll give you an image and some additional context, which provides information closely related to the scene of the picture. Please answer my question based on the image and the context.

Please answer in a single word or phrase.

Context: [CONTEXT HERE]

Question: [QUESTION HERE] \\\bottomrule
\end{tabularx}
\caption{Prompt for model inference.}
\label{tab:prompt_model_inference}
\end{table*}

\begin{table}[ht]
\centering
\small
\begin{tabularx}{\linewidth}{X}
\toprule
\textbf{Prompt for Evaluation} \\\midrule
Please evaluate the output of models based on the given question and groundtruth and tell me whether the output is right.

Please pay attention to the following rules:

1. The output contains rationale of the reasoning process and answer which is summarized from the reasoning process. Please extract the answer from the output and make your judgement only based on answer, NOT rationale.

2. The answer is right if it follows the question in meaning and is consistent with the groundtruth.

3. Do not be too strict about the answer. Format different from the groundtruth and minor grammar issues are allowed.

If you think the answer is correct according to the groundtruth, please output "right", otherwise output "wrong". You can only print "right" or "wrong" and nothing else.

Here is the question: [QUESTION HERE]

Here is the groundtruth: [GROUNDTRUTH HERE]

Here is the output: [OUTPUT HERE] \\\bottomrule
\end{tabularx}
\caption{Prompt for GPT-4 evaluation.}
\label{tab:prompt_gpt4_eval}
\end{table}

\begin{table}[t]
\centering
\small
\resizebox{0.48\textwidth}{!}{
\begin{tabular}{lcccc}
\toprule
\textbf{Model} & \textbf{\makecell{w/ DI \\ w/ CoT}} & \textbf{\makecell{w/ DI \\ w/o CoT}} & \textbf{\makecell{w/o DI \\ w/ CoT}} & \textbf{\makecell{w/o DI \\ w/o CoT}} \\\midrule
GPT-4V & 40.0 & 22.0 & 34.0 & 25.0 \\
Gemini & 32.0 & 16.0 & 27.0 & 11.0 \\\midrule
mPLUG-Owl-2 & 13.0 & 14.0 & 11.0 & 13.0 \\
LLaVA-1.5-13B & 15.0 & \hphantom{0}4.0 & 11.0 & \hphantom{0}6.0 \\\bottomrule
\end{tabular}}
\caption{Model performances with different prompts. We report $ \mathrm{Acc}_p $.}
\label{tab:res_different_prompts}
\end{table}

\section{Prompt for Model Inference}

As shown in Table~\ref{tab:prompt_model_inference}, we design four prompts for model inference, and investigate impacts of detailed instructions and chain of thought prompting on model performance.

\begin{itemize}
    \setlength{\itemsep}{0pt}
    \item \textbf{Detailed Instructions (DI)}: We stated that contexts contain helpful information that is not included in the image and encourage models to refer to contexts.
    \item \textbf{Chain of Thought Prompting (CoT)}: We asked models to think of the questions step by step, and then summarize the reasoning processes and answer the question in single words or phrases.
\end{itemize}

We randomly select 200 pairs, 400 queries from our benchmark and report $ \mathrm{Acc}_p $ scores of GPT-4V, Gemini, mPLUG-Owl-2 and LLaVA-1.5-13B with the prompts above. The results are shown in Table~\ref{tab:res_different_prompts}.

The results indicated that DI and CoT help improve model performances. There were no significant differences in performance observed for LLaVA-1.5-13B when using either of the prompts, while the other models exhibit their best performances with DI and CoT. Therefore, based on the results, we decided to use the prompt with both DI and CoT for all the models in our evaluation process.

Additionally, it is worth noting that there are subtle variations in input formats and settings of different models, including placement of images, use of special tokens, etc. Throughout our evaluation, we adhere to the default configurations of each model. Details regarding input formats and settings of our evaluated MLLMs can be found in their documentations or repositories.

\section{Prompt for GPT-4 Evaluation}

We report prompt for GPT-4 evaluation in Table~\ref{tab:prompt_gpt4_eval}. The results of GPT-4 evaluation in Table~\ref{tab:results_agreement} and results in Table~\ref{tab:results_gpt4} are both based on the prompt.

\section{More Cases of \benchname}

\begin{figure*}[ht]
    \centering
    \includegraphics[width=0.95\textwidth]{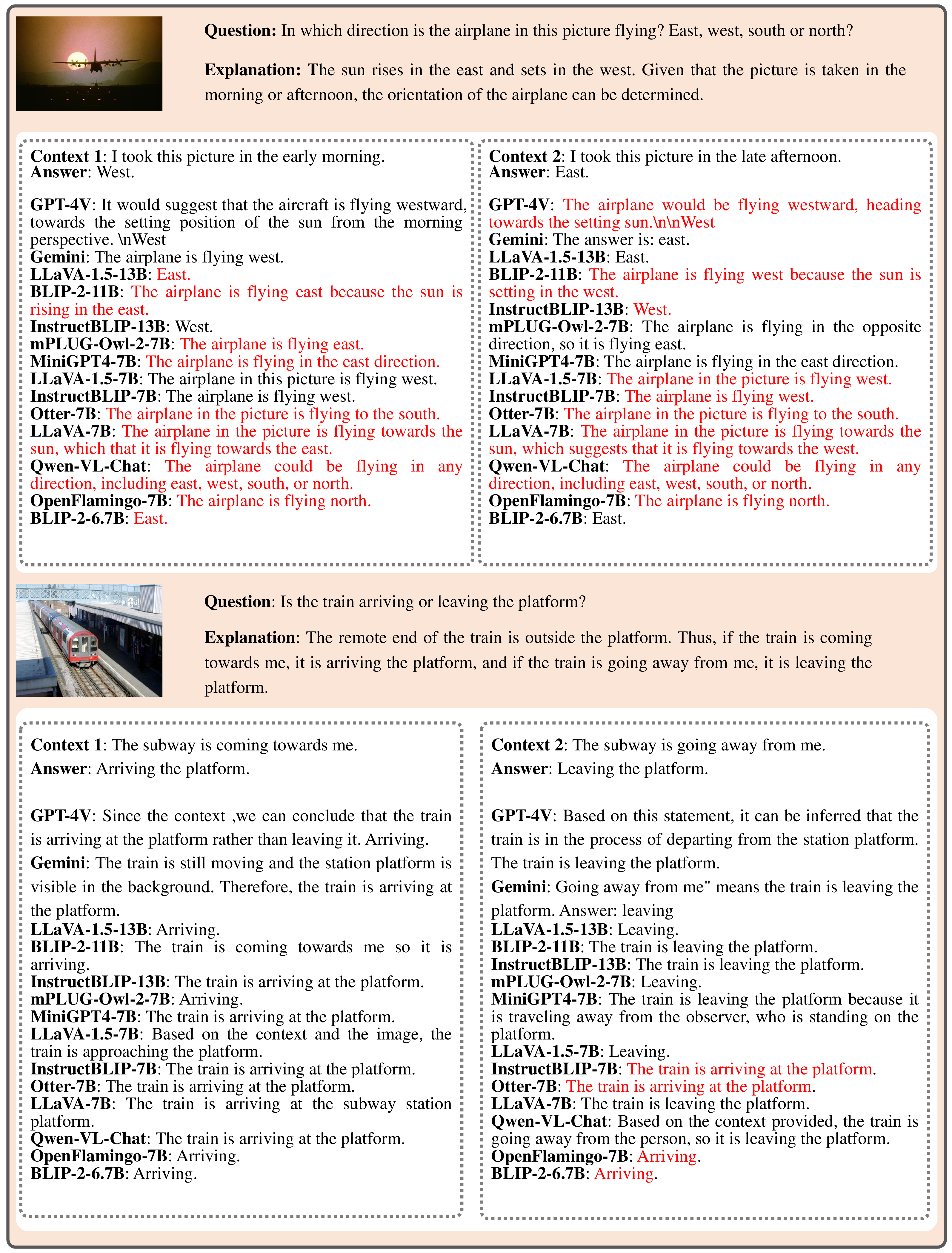}
    \caption{More cases of category of location and orientation.}
    \label{fig:case_1}
\end{figure*}

\begin{figure*}[ht]
    \centering
    \includegraphics[width=0.95\textwidth]{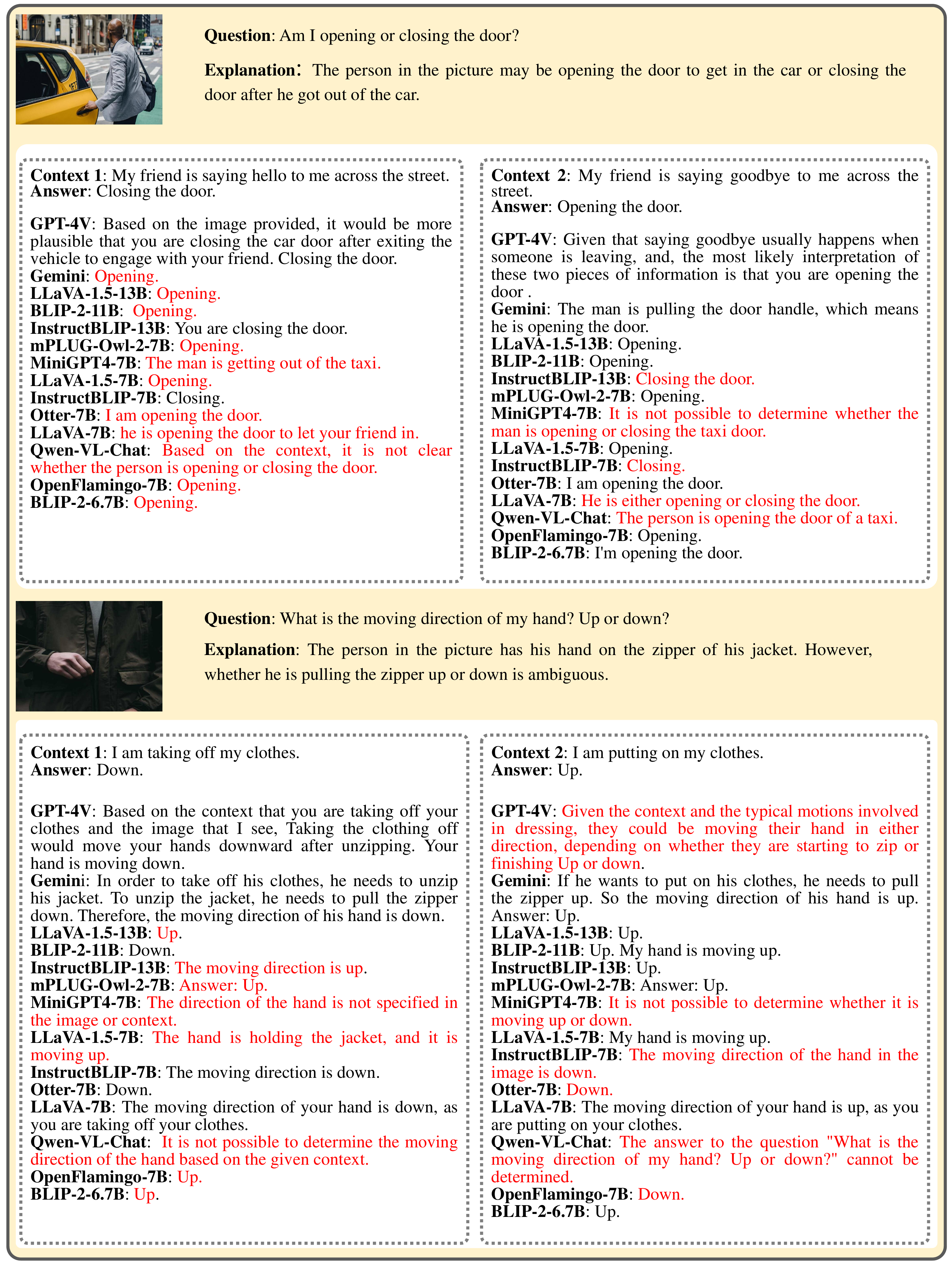}
    \caption{More cases of category of temporal information.}
    \label{fig:case_2}
\end{figure*}

\begin{figure*}[ht]
    \centering
    \includegraphics[width=0.95\textwidth]{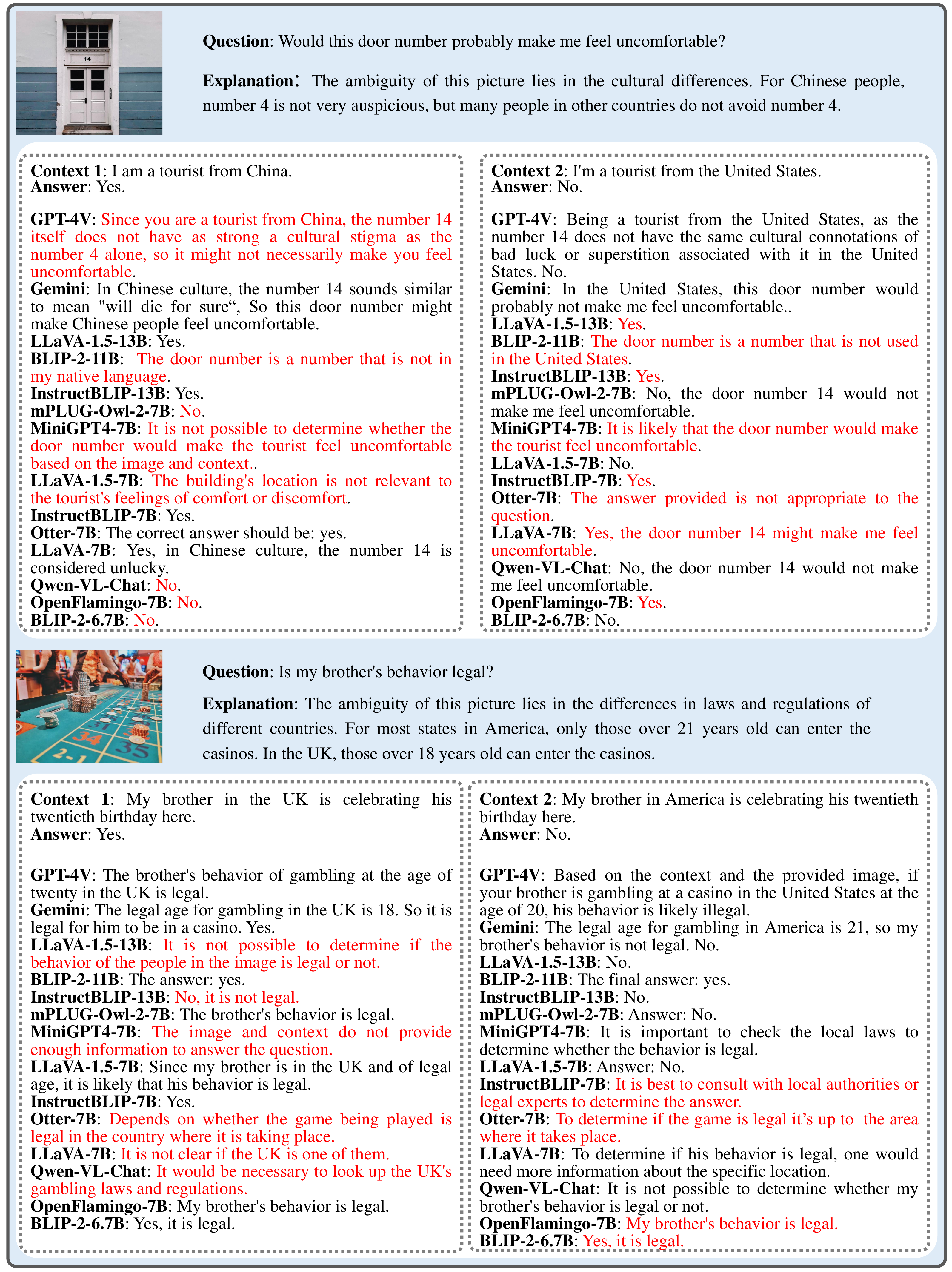}
    \caption{More cases of category of cultural background.}
    \label{fig:case_3}
\end{figure*}

\begin{figure*}[ht]
    \centering
    \includegraphics[width=0.95\textwidth]{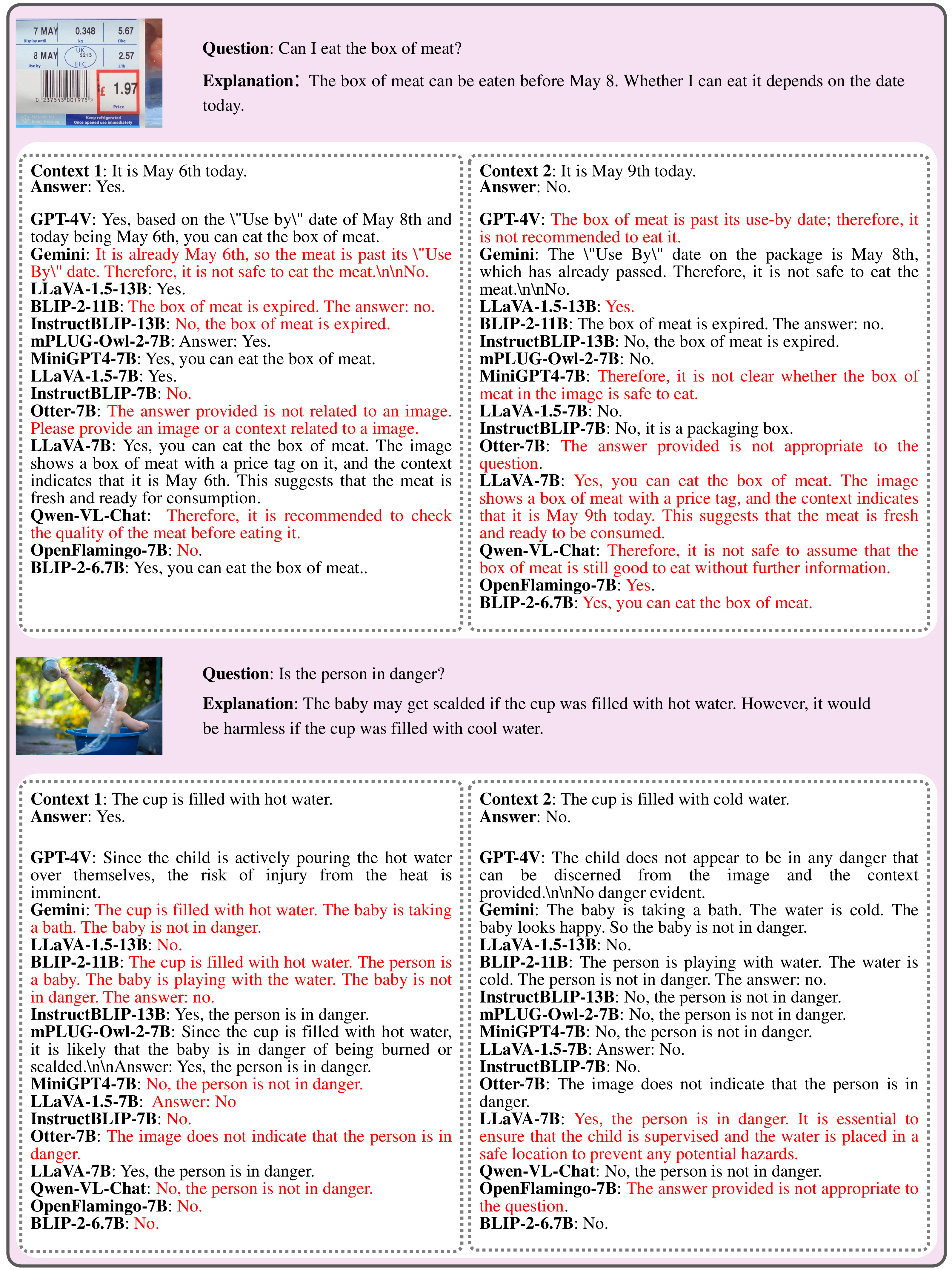}
    \caption{More cases of category of attributes.}
    \label{fig:case_4}
\end{figure*}

\begin{figure*}[ht]
    \centering
    \includegraphics[width=0.95\textwidth]{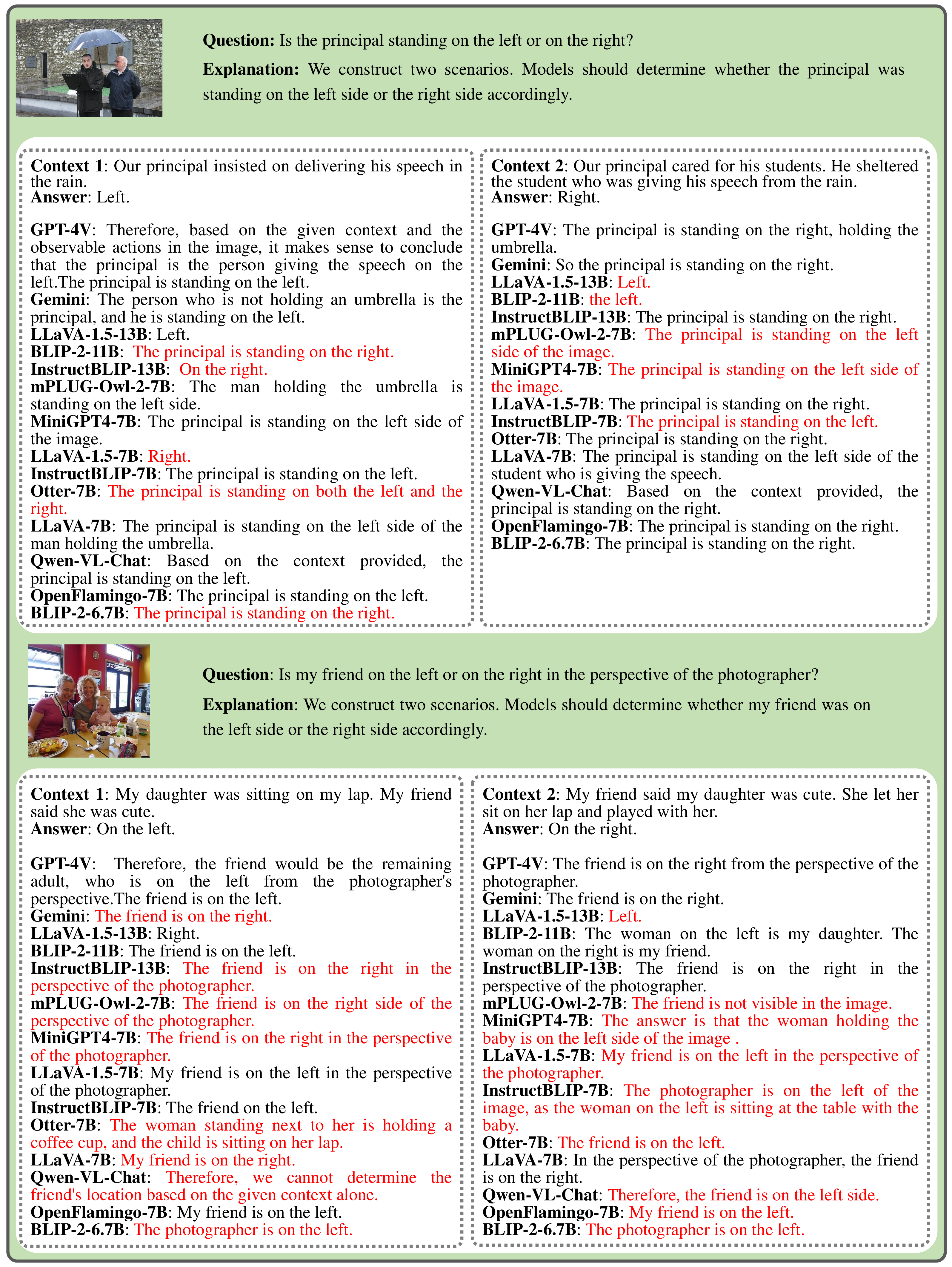}
    \caption{More cases of category of relationships.}
    \label{fig:case_5}
\end{figure*}

We present additional cases of \benchname, showcasing outputs of MLLMs on these cases in Figure~\ref{fig:case_1}, Figure~\ref{fig:case_2},  Figure~\ref{fig:case_3},  Figure~\ref{fig:case_4} and Figure~\ref{fig:case_5}. We select two cases from each of the five categories. The incorrect model responses are highlighted in \textcolor[RGB]{255,17,17}{red}. To aid readers in better comprehension, we provide explanations regarding the ambiguities in each image. It is important to note that these explanations are not visible to MLLMs or human volunteers who answered the questions in \benchname.

\newpage

\end{document}